\pdfoutput=1

\documentclass[11pt]{article}

\usepackage[final]{coling}

\usepackage{times}
\usepackage{latexsym}
\usepackage[T1]{fontenc}

\usepackage[utf8]{inputenc}
\expandafter\def\expandafter\UrlBreaks\expandafter{\UrlBreaks
  \do\a\do\b\do\c\do\d\do\e\do\f\do\g\do\h\do\i\do\j%
  \do\k\do\l\do\m\do\n\do\o\do\p\do\q\do\r\do\s\do\t%
  \do\u\do\v\do\w\do\x\do\y\do\z\do\A\do\B\do\C\do\D%
  \do\E\do\F\do\G\do\H\do\I\do\J\do\K\do\L\do\M\do\N%
  \do\O\do\P\do\Q\do\R\do\S\do\T\do\U\do\V\do\W\do\X%
  \do\Y\do\Z}

\usepackage{soul}
\usepackage[normalem]{ulem}
\usepackage{xcolor}

\usepackage{enumitem}
\setlist{nolistsep}
\usepackage{microtype}

\iffalse 
    \usepackage[final]{proofing}
  \renewcommand\hl[1]{{#1}}  
   {\draftnote{\red{#2}}}
   \newcommand\redHL[1]{}
  \newcommand\todo[1]{}

  \newcommand{\Djame}[1]{}

\else  

\usepackage[final]{proofing}

\newcommand{\Djame}[1]{
\textbf{\textcolor{red}{\hl{Djame: #1}}}
}

\usepackage{titlesec}

\newcommand\red[1]{{{\textcolor{red}{\bf #1}}}}

\let\oldred\red
\renewcommand\red[1]{{ \oldred{{#1}}}}

 \newcommand\redHL[1]{\red{\hl{#1}}}
\let\olddraftnote\draftnote
\renewcommand\draftnote[1]{\olddraftnote{\red{#1}}}

\fi

\usepackage{microtype}
\usepackage{booktabs} 
\usepackage{multirow}
\usepackage{inconsolata}

\usepackage{graphicx}
\usepackage{subcaption}

\usepackage{xspace}
\newcommand{\xlmt}{{\sc XLM-T}\xspace}
\newcommand{\xlmr}{{\sc XLM-R}\xspace}
\newcommand{\multirad}{{\sc Counter}\xspace}
\title{Beyond Dataset Creation: Critical View of Annotation Variation and Bias Probing of a Dataset for  Online Radical Content Detection}

\author{Arij Riabi  \quad Virginie Mouilleron \quad Menel Mahamdi\\  \quad {\bf Wissam Antoun} \quad {\bf Djamé Seddah} \\
      Inria, Paris\\
     \{firstname,lastname\}@inria.fr}

\begin{document}
\maketitle
\begin{abstract}


The proliferation of radical content on online platforms poses significant risks, including inciting violence and spreading extremist ideologies. Despite ongoing research, existing datasets and models often fail to address the complexities of multilingual and diverse data. To bridge this gap, we introduce a publicly available multilingual dataset annotated with radicalization levels, calls for action, and named entities in English, French, and Arabic. This dataset is pseudonymized to protect individual privacy while preserving contextual information. Beyond presenting our \href{https://gitlab.inria.fr/ariabi/counter-dataset-public}{freely available dataset}, we analyze the annotation process, highlighting biases and disagreements among annotators and their implications for model performance. Additionally, we use synthetic data to investigate the influence of socio-demographic traits on annotation patterns and model predictions. Our work offers a comprehensive examination of the challenges and opportunities in building robust datasets for radical content detection, emphasizing the importance of fairness and transparency in model development.

\end{abstract}

\section{Introduction}
Given the current debate on the influence of social media and their lack of moderation, making it a giant echo chamber for all kinds of ideologies, it is an understatement to say that the detection of radical content on online platforms has become an increasingly pressing concern. Indeed, radicalization, often driven by online propaganda, has contributed to recent terror attacks and public violence. For example, the United Kingdom experienced a baffling rise in racially motivated attacks\footnote{\url{https://edition.cnn.com/2024/08/05/uk/uk-far-right-protests-explainer-gbr-intl/index.html}}, whose impact was amplified by the viral spread of many related videos. In this context, online expressions of radicalization pose a unique challenge as they can constitute a rallying point for potentially burgeoning communities and then provide direct access to such communities where extreme opinions can be further intensified \cite{Bowman-Grieve2010,Nouh2019UnderstandingTR,10.1145/2757401.2757408,baele2024}. Beyond the spread of ideas, online extremism can lead to {\em offline} dangers, including violent riots, terrorist attacks and so on \cite{farwell2014,fernandez2021artificial,PELLICANI2023435}. An important point is to note that the rapid spread of information online, mainly through social media, enables extremist groups to disseminate radical content and recruit others to their cause. However, this is often the first step before these groups migrate to encrypted platforms, evading regulation and oversight.
This is why trying to understand the interplay between radicalization process, social network dynamic and human interactions is a crucial challenge. Studies \citep{10.1093/poq/nfw006,doi:10.1126/science.aaa1160,dekock2024jointlymodellingevolutioncommunity} have explored how exposure to varying ideological perspectives online influences individuals,  highlighting the importance of analyzing echo chambers in social media, where the most radicalized and polarized views tend to dominate the discourse \cite{roy2021identifying}. \draftremove{For example, a tweet expressing ``thoughts and prayers'' following a gun attack may subtly reveal the author's political leanings, advocating for stricter gun control rather than offering genuine condolences.}
Few annotated datasets cover radical content from social media \cite{fernandez2021artificial}.

As data quality directly affects model performance and user trust, developing high-quality, consistently labeled data is essential. Our research addresses this challenge by offering a comprehensive analysis of the dataset creation process. We explore how variations in annotations and model training impact radical content detection in NLP. We present \multirad, a novel pseudo-anonymized dataset created to tackle the complexities of radical content detection across multiple languages—English, French, and Arabic-and ideologies, from far-right to Jihadism. We release the dataset with multiple annotations (NER, {\tt Radical Level}, {\tt Call for Action}), annotators disagreement, all the guidelines, and the synthetic data for bias analysis. We seek to understand the interplay between annotation biases, model generalization, and fairness in detecting radical content. First, we analyze how human label variations affect model outcomes, highlighting how it is heavily dependent on the aggregation method and the evaluation. Second, we conduct an in-depth study of the annotations to identify the most suitable experimental settings to improve model performance. Third, we introduce and evaluate synthetic data as a bias analysis tool, simulating socio-demographic attribute's influence on model predictions.
Our results highlight the complexities in detecting radical content, especially given the inherent subjectivity in human annotations and the sociodemographic variations that influence data and model outcomes.

\section{Online Radical Content Detection}

Radical content can be defined as a signal used by an individual or group of individuals to express a radical perspective in opposition to a political, social, or religious system, and adopting a radical discourse could be followed by a progressive shift in social behavior, resulting in violence and even serious undermining of public safety \cite{fink2014understanding}.
The definition of radicalization itself is fluid, evolving with the phenomena and associated events, making it difficult for detection algorithms to maintain effectiveness as the associated behaviors and language evolve \cite{Berjawi2023,Schmid2016}. This ongoing evolution complicates the definition of radicalization and diminishes the efficiency of detection models as the language and behaviors indicative of radicalization shift over time. 

\paragraph{NLP for radical content detection}
NLP methods show potential for detecting radicalization but require further exploration, as indicated by literature \cite{9300086,de-kock-hovy-2024-investigating}. Analyzing radicalization mechanisms using NLP techniques has been mostly done in a supervised learning setting for different steps like propaganda, recruitment, networking, data manipulation, and disinformation \cite{hung2019,Torregrosa2021ASO,Aldera2021OnlineED}. However, existing datasets used for radicalization detection tend to have a narrow focus, often focusing on specific behaviors within particular extremist communities, thus lacking a broader perspective on radicalization across different groups \cite{hartung-etal-2017-ranking,alatawi2021}. The quality and availability of training and evaluation datasets are significant constraints in radicalization detection, and large datasets often suffer from biases and inadequate quality checks \cite{Gaikwad2021OnlineED}. Many are gathered using simplistic rules, such as identifying users who employ specific lexicons or share certain content \cite{LaraCabrera2017MeasuringTR,Fernandez2018}. These rules often rely on unverified assumptions, introducing noise and reducing dataset quality. Furthermore, when human annotators evaluate data, only a small subset of content is manually verified \cite{ashcroft2015detecting,agarwal2015using,10.1155/2023/4563145}, with annotations often performed through crowdsourcing platforms rather than domain experts, adding additional biases. We found that the literature on radicalization detection in NLP focuses on two primary objectives: detecting online radical content and identifying radicalized users and communities. 
\paragraph{Investigating radicalized users} allows researchers to detect early-stage radicalization by analyzing behavioral changes over time \cite{barachi,sakketou-etal-2022-investigating}. However, challenges such as content deletion, account changes, and cross-posting complicate this task. On this topic, \citet{de-kock-hovy-2024-investigating} emphasize the lack of research in NLP and propose a semi-supervised solution to bypass a "potentially biased human annotation step." They focus on sociolinguistic indicators like hostility, longevity, and social connectivity, using a lexicon for hostility and radical ideologies \cite{10.1145/3292522.3326045}. 


\paragraph{Detecting radical content} can also serve as an indicator for identifying potential radicalized users. Especially after linguistic analysis of online communication from high-risk groups revealed the existence of linguistic characteristics that distinguish them from general discourse \cite{Winter_Neumannt_2021,Mueller2022}. Some datasets have been proposed for radical content detection, with the ISIS dataset from Twitter being the most common \cite{desmedt2018automaticdetectiononlinejihadist}. A significant challenge in building these datasets is defining adequate annotation schemes. Most datasets still treat radicalization as a binary state \cite{agarwal2015using}, which oversimplifies its complex and gradual nature. Some works have attempted to refine this approach by differentiating between hostile and irrelevant content \cite{ashcroft2015detecting,abrar2019framework,kaur2019detectingradicaltextonline} or by categorizing content into propaganda, radicalization, and recruitment \cite{10.1155/2023/4563145}. While most of the research focuses on English with a US-centered perspective, few works focusing on other languages can be found, such as Indonesian \cite{Miranda2020} and Arabic \cite{Aldera2021}, particularly in the context of jihadism. As far as we know, no large-scale multilingual works have been conducted in this domain. This is why to fully address the complexity of radical content across diverse contexts, a multilingual dataset with rich annotations from various sources is essential. This can enable the study of radicalization across cultural and linguistic boundaries, providing more nuanced insights into the detection of radical content.
\section{\textsc{Counter}: Radical Content Dataset}
\subsection{Data collection}
The dataset includes English, French, and Arabic posts from various sources that can be split between social media (Facebook, Twitter), platforms (Telegram), and forums either public, such as Reddit or banned from search engines (4chan) available via special software such as Tor (a set of tools that enables anonymous communication) enabling access to what is often referred as ``Darkweb''. The contractor carried out the data collection using a list of keywords inspired by relevant geopolitical events. The content posts cover two main ideologies (Jihadism and Far-right), each with different levels of sub-ideologies. Content that cannot be grouped in the previous two categories is put in a third category, which includes posts that do not directly align with Jihadist or Far-right ideologies but still exhibit radical tendencies. The category distribution varies by language, with Far-right dominant in English and French and Jihadism in Arabic. All the meta-data with the posts were kept when available, including the extraction date, post date, and interaction information. Images and video links were also collected 
(Cf. Appendix \ref{sec:statement} for a detailed data statement \cite{bender-friedman-2018-data}).
\begin{table}[h!]
\centering
\footnotesize
\begin{tabular}{lllll}
\toprule
& Arabic & English & French  \\
\midrule
\#sentences & 2499 & 2650 & 2650 \\
\#tokens &  168.48K & 100.73K & 87.58K  \\
Avg length & 67.42 & 38.01 & 33.06 \\
\#NER entities & 6579 & 6651 & 4884 \\
\# anonymized sents& 1500 & 2650 & 2650  \\ 
\# anonymized entities& 130 & 1615 & 649  &  \\ 
\bottomrule
\end{tabular}
  \caption{Dataset Overview.}
  \label{tab:dataset_overview}
\end{table}

\subsection{Prescriptive Annotation}
Annotating radicalized data requires ``experts'' in the domain. Annotators must have native-like knowledge of the target language and its linguistic and cultural aspects (Ex, recognizing puns based on specific cultural references like Klan Chowder). Therefore, a contractor with expertise in this task created the first dataset version. This version follows a prescriptive approach to annotation \cite{rottger-etal-2022-two}, discouraging subjectivity and aiming for consistency by adhering to predefined guidelines. The contractor received the specifications and the needs to produce a dataset to train a detection model for the platform, which is supposed to be used to facilitate online moderation of radicalization content. The annotations are multi-label and multi-class. The main label is {\tt Call for Action}, with five predefined levels based on the degree to which it motivates specific actions, ranging from ``negative'' to ``very high''. We also have annotations for {\tt Radicalization Level} with six levels covering ``Negative'' or ``Neutral'' content, ``Expression of Radical Views'', ``Using Radical Propaganda'', ``Associated with Radical Groups'', ``Dehumanizing the Other'' and ``Call for Action against others''. (See Appendix \ref{subsec:class_description} for more details).  
\subsection{Descriptive Annotation}
We lacked detailed information about the contractor's annotation process, which is crucial for investigating biases in subjective tasks. Therefore, we added double annotations for {\tt Call for Action} Classification and {\tt Radicalization Level} for English and French\footnote{Due to the potential psychological impact of annotating radical content, psychological support was made available to annotators to ensure their well-being throughout the process.}. We adopted the descriptive paradigm \cite{rottger-etal-2022-two} for those annotations. We gave the annotators the contractor's task description and discussed the task and possible use cases. However, we relied on the lack of details for the annotation to encourage annotators' subjectivity. We wanted the annotations to represent a larger range of beliefs to extract related insights since correlations have been shown between socio-demographic factors and annotations for different tasks such as sentiment analysis \cite{diaz2018} and hate speech \cite{waseem-2016-racist,sap-etal-2022-annotators}. We recruited two trained linguists with different socio-demographic profiles.
Both female annotators are between [25-30] and [40-45], have a master's degree as their highest completed education, are native French speakers, and are fluent in English. Re-annotation guidelines were developed to provide more detailed instructions and accommodate the annotators' interpretations.
After defining the guidelines, additional uncertainties were addressed at different times during the process. They double-annotated the French dataset and a large sample of the English dataset. When uncertain, they were encouraged to select the most appropriate class while we tracked those cases.

We report in Table \ref{tab:cohen_kappa_agreement} in Appendix \ref{sec:MultiRad_dataset}  the inter-annotator agreement for {\tt Radicalization Level} and {\tt Call for Action}. The French dataset shows higher agreement, with moderate agreement for {\tt Radicalization Level} at Fleiss' Kappa 0.50 and fair agreement for {\tt Call for Action} at 0.43. In contrast, the English dataset shows lower agreement, with fair agreement for {\tt Radicalization Level} at Fleiss' Kappa 0.26 and slight agreement for {\tt Call for Action} at 0.13.
\paragraph{Pseudononymization and NER annotations} were performed simultaneously by the annotators. The main goal was to preserve all semantic properties that can be extracted from the dataset. Our approach ensures the protection of sensitive information without losing critical data, which facilitates sharing the dataset for research purposes. We kept well-known events and public figures non-anonymized to leverage the model's embedded knowledge and maintain alignments within the text. We explain the detailed pipeline in \citet{riabi-etal-2024-cloaked}. We analyzed the effect of the pseudonymization and found that training on both datasets gave comparable results (Appendix \ref{subsec:anonymization_effect}).

\subsection{Synthetic Data for Bias Analysis}
While investigating how socio-demographic traits influenced model decisions, we faced challenges in directly extracting this information from the posts in our dataset, resulting in substantial time and resource constraints. To address this, we adopted the recent trend of using large generative models to create examples with socio-demographic information \cite{durmus2024measuringrepresentationsubjectiveglobal,pmlr-v202-aher23a}, which reduces privacy risks and annotations cost \cite{Argyle_Busby_Fulda_Gubler_Rytting_Wingate_2023}. This technique, referred to as ``persona prompting'', is often used to simulate survey participants \cite{jiang-etal-2022-communitylm,simmons2023largelanguagemodelssubpopulation} or annotators \cite{lee-etal-2023-large,hu-collier-2024-quantifying}. The effectiveness of such techniques remains debatable among researchers \cite{pmlr-v202-santurkar23a,bisbee_clinton_dorff_kenkel_larson_2023,grossmann2023}, but promising results have been demonstrated in several cases \cite{simmons-savinov-2024-assessing,jiang-etal-2024-personallm}. While many works using generative models use in-context learning to guide the generation of examples \cite{simmons-savinov-2024-assessing}, it was unsuitable for our case due to the lack of annotated examples with relevant socio-demographic variables, and using in-context learning would have influenced the model with inaccurate or incoherent profiles. Instead, we opted for a zero-shot prompting setup.
Our approach is based on creating user profiles that include socio-demographic variables such as age and gender, income, education level, and more ( See Appendix \ref{sec:prompt} for a complete list of the variables and a prompt example). We tried to include as many protected characteristics as possible \footnote{Protected characteristics correspond to attributes of people that anti-discrimination law mentions explicitly Cf. \href{https://commission.europa.eu/aid-development-cooperation-fundamental-rights/your-rights-eu/know-your-rights/equality/non- discrimination_en}{link}} as most studies focus on gender and ``race''\cite{sotnikova-etal-2021-analyzing}. These profiles were used to prompt an uncensored LLM called \textit{Wizard-Vicuna-13B-Uncensored}~\cite{WizardVicuna13B}, which was trained on a dataset with its alignment guardrails intentionally removed. The model was created by combining the WizardLM~\cite{xu2024wizardlm} approach to training dataset generation with Vicuna's~\cite{vicuna2023} multi-turn conversational data, allowing for more open and flexible interactions. The LLM was used to generate posts annotated for {\tt Radicalization levels} and {\tt Calls to action} by the annotators who had previously re-annotated the original dataset. There was moderate agreement between annotators on a sample of 300 examples, with Cohen's Kappa scores of 0.51 and 0.40 for English and 0.54 and 0.47 for French on {\tt Radicalization Level} and {\tt Call for Action}, respectively. The process involves generating both ``base profiles'' and ``variation profiles'' by altering a few variables to maintain consistency while ensuring the authenticity of the generated content. Profiles were also created by taking inspiration from real profiles in the \multirad dataset. The generated posts generally reflect socio-demographic variables but sometimes rely heavily on stereotypical keywords according to the annotators. The model accurately incorporates most variables but struggles with differentiating based on age and does not always consider the register. 

\section{Results}

\textbf{Experimental setting.} We use the MaChAmp v0.2 toolkit \cite{van-der-goot-etal-2021-massive}, a framework that allows the implementation of different transformers-based tasks and supports single-task and multi-task learning. In the multi-task setting, the encoder is shared between the tasks, which are jointly fine-tuned during training, while we have a different decoder per task.
We split our datasets approximately to 70\% train, 10\% validation, and 20\% test for each language with stratification across ideologies, {\tt Call for Action}, and {\tt Radicalization level} (More Details in Appendix \ref{subsec:splitting}). We report the average Macro-F1 over five seeds on the test set and use the validation set to pick the best checkpoint.

\textbf{Baseline.} We fine-tune \xlmt \cite{barbieri-etal-2022-xlm}, an \xlmr \cite{conneau-etal-2020-unsupervised} model, that has been fine-tuned for the MLM task on 200 million tweets in more than 30 languages, which makes it more adapted for social media data. We report in Table \ref{tab:results_feats} the results for our main task, multi-class classification of {\tt Call for Action}. The baseline results show that \xlmt performs reasonably well across all languages, with Macro-F1 scores ranging from 59.41 for Arabic to 65.65 for French.
\begin{table}[htb!]
\centering
\footnotesize
\begin{tabular}{@{}lccc@{}}\toprule
 & {en} & {fr}& {ar}\\\midrule
 Baseline  &64.63\scriptsize{($\pm$2.0)} & 65.65\scriptsize{($\pm$1.8)} & 59.41\scriptsize{($\pm$1.3)}\\
 + Radical level &  64.82\scriptsize{($\pm$2.5)} & 63.91\scriptsize{($\pm$6.5)} & 58.98\scriptsize{($\pm$2.4} \\
 + Ideology pred &  65.10\scriptsize{($\pm$1.7)} & 65.56\scriptsize{($\pm$8.6)}  &57.84\scriptsize{($\pm$2.6)}\\
 + NER ID &  64.64\scriptsize{($\pm$1.3)} & 61.98\scriptsize{($\pm$3.9)} & - \\
 + NER OOD &  66.47\scriptsize{($\pm$2.5)} & 63.74\scriptsize{($\pm$4.5)} & 58.37\scriptsize{($\pm$1.5)} \\
 \midrule
 {{\sc Multi-same}\xspace} &  63.98\scriptsize{($\pm$1.9)} & 60.87\scriptsize{($\pm$4.7)} & 56.85\scriptsize{($\pm$2.3)} \\
  {{\sc Multi-diff}\xspace} &  \textbf{66.65}\scriptsize{($\pm$4.3)} & \textbf{68.77}\scriptsize{($\pm$3.2)} & \textbf{59.48}\scriptsize{($\pm$4.3)} \\
\bottomrule
\end{tabular}
\caption{Macro-F1 of {\tt Call for Action} on the test set for \xlmt.}
\label{tab:results_feats}
\end{table}

\textbf{Does Adding Additional Features Improve the Performance?}
We examine whether incorporating additional features or tasks could improve model performance. Following \citet{montariol-etal-2022-multilingual}, we perform multi-task training to assess whether these supplementary tasks provided helpful context (cf. Table \ref{tab:results_feats}). 

Adding \textbf{Radicalization Level prediction} as an auxiliary task did not improve performance substantially, with slight variations across languages. In particular, performance in French dropped notably, suggesting that this feature may introduce noise rather than aid in prediction. This observation is coherent with the annotator's observations about the variations across datasets regarding the classification scale as they noted that the tone in the English dataset is far more crude, violent, and derogatory than in the French dataset. 

When incorporating \textbf{Ideology Prediction} as an auxiliary task, we observed minor improvements in English, while performance in French and Arabic remained relatively stable or declined slightly. This suggests that the benefit of ideology prediction may depend on language-specific features or underlying data distribution. 

Next, we added NER both within-domain (ID) (Arabic excluded as only 1500 out of 2500 examples were annotated.) and out-of-domain (OOD). For OOD data, we used CoNLL-2003 \cite{tjong-kim-sang-de-meulder-2003-introduction} for English, ANERcorp \cite{obeid-etal-2020-camel} for Arabic, and FTB-NER \cite{ortiz-suarez-etal-2020-establishing} for French. The within-domain NER did not improve results much, with performance even dropping in French. However, NER OOD showed more promise, especially in English, where we observed the most significant improvement, indicating that OOD entities might offer valuable contextual signals for the model.

Finally, the \textbf{multilingual} experiments demonstrated that using separate classification layers for each language ({{\sc Multi-diff}\xspace}) led to the best performance across all languages. This setup outperformed the single-classifier approach ({{\sc Multi-same}\xspace}), suggesting that handling the nuances of each language separately is more effective than sharing a classifier across all languages.
\section{Discussion}
We recognize the importance of providing a clear overview of what to expect from models trained on our dataset. In this discussion, we focus on three key aspects. First, we examine how human label variations and annotator disagreements impact model performances. Second, we address fairness concerns, exploring how model predictions may disproportionately affect different demographic groups. Finally, we compare multi-class classification and regression approaches to better capture the nuances of radical content detection.
\begin{table}[htb!]
\centering
\footnotesize
\begin{tabular}{@{}lccc@{}}\toprule
&\multicolumn{3}{c}{Test} \\
\cmidrule(lr){2-4}
{Train} & {Contractor} & {MACE}& {Majority}\\\midrule
 Contractor  &\textbf{65.65}\scriptsize{($\pm$1.8)} & 50.25\scriptsize{($\pm$1.7)} & 52.57\scriptsize{($\pm$1.8)}\\
  Mace  &53.21\scriptsize{($\pm$1.7)} & \textbf{57.92}\scriptsize{($\pm$1.4)}& \textbf{57.63}\scriptsize{($\pm$0.7)}\\
 Majority  &53.53\scriptsize{($\pm$1.4)} & 55.59\scriptsize{($\pm$1.1)} & 56.73\scriptsize{($\pm$2.0)}\\
Repeated-lab  &56.46\scriptsize{($\pm$1.6)} & 56.02\scriptsize{($\pm$0.8)} & 56.91\scriptsize{($\pm$1.6)}\\
Annot-classifier  &58.31\scriptsize{($\pm$4.8)} & 55.12\scriptsize{($\pm$1.2)} & 56.96\scriptsize{($\pm$1.3)}\\

\bottomrule
\end{tabular}
\caption{Macro-F1 results on the test set for human label variation analysis for French set and \xlmt model.}\label{tab:results_variations}
\end{table}

\subsection{How about Human Label Variations?}
Annotator disagreement has been observed for different tasks in literature \cite{jamison-gurevych-2015-noise,larimore-etal-2021-reconsidering,peng-etal-2024-different}. A relatively recent line of work called ``perspectivist paradigm'' \cite{fleisig-etal-2024-perspectivist} investigates the best way to handle the variability in annotations \cite{uma2022,barrett-etal-2024-humans}. This variability can be caused by several reasons, such as task ambiguity, unclear guidelines, annotator expertise, data complexity, and the most challenging reason, subjectivity \cite{sandri-etal-2023-dont}. Therefore, we leverage our multiple annotations to explore how annotator disagreements impact model performance and whether incorporating diverse labels can mitigate biases. We adopt the concept of \textit{Human label variation} as defined by \citet{plank-2022-problem}, considering societal biases and interpretative disparities as the primary sources of disagreement in our task.\footnote{We note that annotator disagreements may also arise from annotation errors or plausible variations, as discussed in \cite{weber-genzel-etal-2024-varierr}, but we leave the distinction between these cases to future work.} Multi-annotator techniques help measure uncertainty when annotations are inconclusive \cite{davani-etal-2022-dealing}. This process is critical in the context of radicalization detection, where factors like ethnicity, gender, and age influence how different communities perceive radicalizing content, shaped by their cultural and social backgrounds \cite{vijayaraghavan2021interpretable,fleisig-etal-2023-majority}. Varying interpretations of radical content may also stem from subjective perceptions, particularly when language includes coded messages or ideological allusions \cite{lee-etal-2024-exploring-cross}. Moreover, annotators' thresholds for identifying dangerous material can differ, further influencing their conclusions \cite{sap-etal-2022-annotators}.
We do not seek the best aggregation method, but we want to show the variability between the approaches and encourage the choice of the adequate option depending on the use of the model. We test four approaches of annotation aggregations: 
\begin{itemize}
    \item Inspired by\cite{plaza-del-arco-etal-2024-wisdom}, we use \textbf{MACE} \cite{hovy-etal-2013-learning} aggregated labels, a Bayesian annotation tool that calculates two scores: the most likely label and the competence (reliability) of each annotator (i.e., the likelihood that an annotator selects the "true" label based on their expertise rather than speculating on one).
MACE operates on unlabeled data and infers both variables using variational Bayesian inference. The aggregated MACE labels are usually more accurate than majority voting.
    \item \textbf{Repeated Labeling}: Treats each annotation as a separate instance. It captures the full range of crowd opinions by treating each label as a separate learning signal.
    \item \textbf{Majority} aggregated the annotations by taking the most frequent label and choosing randomly in case of a tie.
    \item \textbf{Annot-classifier} A single classification head models each annotator. The three predictions are aggregated using a majority vote.
\end{itemize}
\begin{figure*}[htb!]
    \centering
    \includegraphics[width=1\textwidth]{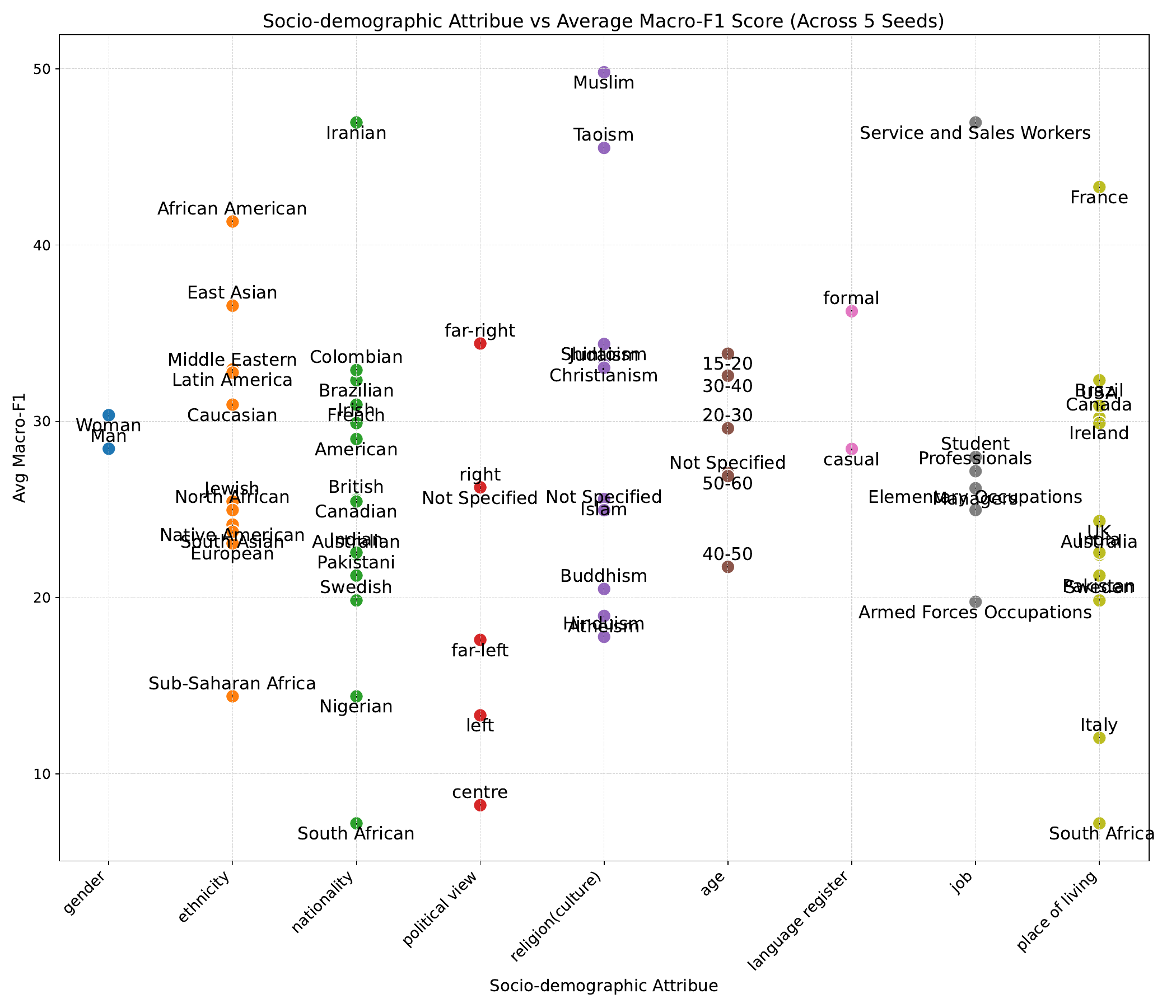}
    \caption{Average Macro-F1 variations for the attributes for \xlmt model for the synthetic English set.}
    \label{fig:bias_f1}
\end{figure*}

We focus on French for this analysis as we have  3 annotations for all examples. We report the results in Table \ref{tab:results_variations}, considering three gold labels: expert's annotations, the majority vote, and MACE.

As expected, the models trained on the corresponding gold standard labels achieved the best results for each test set. Except for the Majority, the best results on majority-labeled test data were from the model trained on MACE aggregation. This highlights MACE's capacity to effectively model variations and capture the underlying consensus among multiple annotators more accurately than majority voting. The drop in performance when models were tested on different gold sets underscores the distinct perspectives each annotation method introduces. For instance, models trained on expert labels showed a notable decrease in performance when tested on MACE and majority labels, suggesting that it offers a more uniform and possibly stricter interpretation than the variability captured by methods like MACE and majority voting.

Cohen's kappa analysis further supports these findings, with high agreement between MACE and majority labels (0.89) and lower agreement between Contractor and MACE (0.70). These results underscore the importance of choosing the appropriate label aggregation method, depending on whether the model needs to align more closely with expert consensus or capture broader interpretations.

\subsection{Bias Analysis: What can we infer?}
We compared the average Macro-F1 score per value for each socio-demographic attribute using our generated dataset to assess model bias. In addition to \xlmt we also trained two other models, \xlmr and {\sc mBert}, on both French and English datasets. The results of these models are reported in Appendix~\ref{subsec:additional_results}, showing that \xlmt outperforms both \xlmr and {\sc mBert} in the English and French test sets. As expected, performance in synthetic data was lower than in the standard test set due to differences in distribution between the training data and the generated dataset. The purpose of this experiment was not to maximize performance, but to diagnose any systematic errors correlated with the attributes. Figure\ref{fig:bias_f1} shows the plot of average Macro-F1 scores across  categories for each attribute for \xlmt (we report the plots for {\sc mBert} and \xlmr in the appendix). 

\paragraph{Sociodemographic biases impact performance}  We observed significant differences across attributes such as nationality, ethnicity, political views, and religion. For instance, all models displayed \textbf{substantial performance variation across political views and ethnicities, indicating a potential bias in how certain groups are represented or classified}. While \xlmt showed the highest overall performance, it exhibited larger disparities between categories than \xlmr and {\sc mBert}. In particular, \xlmt had more pronounced political views and nationality variations, while \xlmr and {\sc mBert} displayed slightly more balanced results but lower overall performance. {\bf This suggests that while \xlmt captures certain patterns better, it may amplify biases in specific demographic categories.} This is likely due to \xlmt being trained on social media data such as tweets, which tend to reflect more polarized and informal language, compared to \xlmr and {\sc mBert}, which were trained on broader corpora like Wikipedia and books, containing more formal and neutral text( 
See Figure \ref{fig:confusion_generated_data} in Appendix \ref{sec:synthetic} for the results of all models on the synthetic dataset).

Interestingly, for the French data, the differences between the three models \xlmt, \xlmr and {\sc mBert} are less pronounced than in English, with more consistent performance across socio-demographic attributes. However, the distribution of the gender attribute \footnote{The data lacks fine-grained classifications of gender identities, focusing only on male and female categories. We do not consider this classification to fully represent gender identities and use it solely for analysis within this specific subset.} shows a larger variation in French results compared to what we observed for English. This could be attributed to the fact that detecting the gender of a user from French texts is generally easier than from English, likely due to many gender-sensitive morpho-syntactic markers in French (Cf. Figure \ref{fig:macro-f1_bias_models_english} and Figure \ref{fig:macro-f1_bias_models_french} for the other plots).


\begin{table}[htb!]
\centering
\footnotesize
\begin{tabular}{@{}lccccc@{}}\toprule
& & \multicolumn{2}{c}{Demographic Parity} & \multicolumn{2}{c}{Equalized odds}\\
\cmidrule(lr){3-4}
			\cmidrule(l){5-6}
{\em ~~~~Attribute}&   & en & fr&en & fr\\
\midrule
\multicolumn{2}{l}{ Place of living }& 0.38 &0.55 &0.61 & 0.62\\
\multicolumn{2}{l}{ Ethnicity }& 0.46 &0.37 & 0.68 & 0.35\\
\multicolumn{2}{l}{ Religion} & 0.32 &0.24 & 0.57 &0.31 \\
\multicolumn{2}{l}{Political view} & 0.24 &0.21 &0.23 &0.24\\
\multicolumn{2}{l}{Age}& 0.21 &0.48 & 0.23& 0.56\\
\multicolumn{2}{l}{Gender}& 0.06 &0.17 &0.05 &0.20 \\
\multicolumn{2}{l}{Job}&0.24  &0.60 &0.41 &0.62 \\
\multicolumn{2}{l}{Nationality}& 0.45 & 0.27&0.66 &0.32 \\
\bottomrule
\end{tabular}
\caption{Demographic Parity Difference and Equalized Odds Difference for \xlmt on synthetic data.}
\label{tab:results_fairness}
\end{table}

\paragraph{Challenges in assessing bias}  Metrics to evaluate bias in the dataset or model decisions are very challenging as they depend on the used aggregation and the task definition \cite{bias2019} "The choice of metrics shapes a research study take-aways". Most metrics quantify the extent to which an algorithm treats people differently and the extent to which an algorithm impacts different people differently. The metrics that are commonly applied assume that there are two outcomes: a favorable one and an unfavorable one. We have a multi-class classification, so we considered class zero as the positive outcome. We use the {\bf demographic parity difference}, 
also called disparate impact; it measures the ratio of favorable outcomes between different groups to assess whether the model treats different groups equally. A demographic parity difference of 0 means that all groups have the same selection rate, which refers to the proportion of individuals in each group who receive a positive outcome. We also use {\bf equalized odds difference}, which seeks that the predictions made by the 
model have equal true positive and false positive rates, regardless of the membership in sensitive groups (Cf. Section \ref{sec:metrics_def} in the Appendix for detailed definitions of both metrics).\footnote{We use the \href{https://fairlearn.org/main/user_guide/assessment/common_fairness_metrics.html}{Fairlearn} library to compute these two metrics}

Results in Table \ref{tab:results_fairness} show that \textbf{overall Demographic Parity is generally smaller than Equalized Odds}, indicating that while the model may be relatively fair in terms of selection rates, it struggles more with ensuring consistency in True Positive Rates  and False Positive Rates  across different groups. The largest disparities in Demographic Parity are observed for  attributes like {\em Place of Living} and {\em Job}, particularly in the French dataset, where the selection rates across groups differ more significantly than in the English dataset. For Equalized Odds, substantial disparities are observed again for \textit{Place of Living}, \textit{Ethnicity}, and \textit{Nationality}, particularly in English. 

Our results imply that the model's predictive performance is less balanced across these groups, with larger differences in accuracy and error rates. Interestingly, \textit{Political View} and \textit{Gender} show the smallest differences in both metrics, suggesting more consistent treatment for these attributes. 

Language-wise, the model generally shows more fairness challenges in French, especially for demographic categories like \textit{Place of Living} and \textit{Job}, with larger disparities compared to English.
\subsection{Multi-Class Classification or Regression?} 

We aim to investigate the complexity of detecting {\tt Call for Action}, focusing on whether multi-class classification or regression yields better performance and evaluation reliability. We trained a regression model using Mean Absolute Error, which measures the average distance between the predicted values and the true labels. To compare this regression approach directly with a classification model, we rounded the regression predictions to the nearest integer, allowing us to calculate the F1 score for both models. Although our classes are \textbf{discrete}, ranging from 0 to 4, they can also be viewed as a continuous spectrum where \textbf{errors between adjacent classes} are \textbf{less severe than between more distant classes}. This suggests that a regression approach, which inherently \textbf{accounts for such gradations}, might be more suitable. However, classification allows the model to learn \textbf{more complex and non-linear patterns} specific to each class, which could capture subtle, class-specific nuances that a regression model might oversimplify. For example, the boundaries between classes 2 and 3 could involve different features or relationships than those between classes 1 and 2, which a classification model is better equipped to learn. 
\begin{table}[htb!]
\centering
\footnotesize
\begin{tabular}{@{}cllll@{}}\toprule
\multirow{2}{*}{Lang}& \multicolumn{2}{c}{Classification} & \multicolumn{2}{c}{Regression}\\
\cmidrule(lr){2-3}
			\cmidrule(l){4-5}
   & Macro-F1 & Spearman&Macro-F1 & Spearman\\
\midrule
 en & \textbf{64.63\scriptsize{($\pm$2.0)}} & 0.75\scriptsize{($\pm$0.03)} &56.21\scriptsize{($\pm$1.5)} & \textbf{0.78\scriptsize{($\pm$0.01)}}\\
 fr&\textbf{65.65\scriptsize{($\pm$1.8)}}   & 0.71\scriptsize{($\pm$0.03)}& 53.15\scriptsize{($\pm$3.8)} & \textbf{0.73\scriptsize{($\pm$0.02)}}\\
  ar&\textbf{59.41\scriptsize{($\pm$1.3)}  } & 0.58\scriptsize{($\pm$0.02)}& 49.46\scriptsize{($\pm$1.8)} & \textbf{0.63\scriptsize{($\pm$0.02)}}\\

\bottomrule
\end{tabular}
\caption{XLM-T's Macro-F1 results trained for {\tt Call for Action} as multi-class classification and  regression.}
\label{tab:results_regression}
\end{table}

We calculated the Spearman correlation coefficient for both models. Spearman correlation is particularly relevant here as it measures the strength and direction of the monotonic relationship between the predicted and actual class rankings. This metric is well-suited for our task because it evaluates how well the models preserve the ordinal nature of the classes, regardless of the predicted values. The results in table \ref{tab:results_regression} show a trade-off between classification and regression models. The classification models achieve higher Macro-F1 scores for all three languages than regression, indicating better accuracy in discrete class prediction. However, regression models slightly outperform in preserving the ordinal structure, as evidenced by higher Spearman correlations, with the biggest improvement for Arabic. The low standard deviations suggest consistent performance across seeds for regression.
\begin{figure}[htb!]
    \centering
    \textbf{Classification}

    \begin{minipage}[b]{0.38\linewidth}
        \centering
        \includegraphics[width=\linewidth]{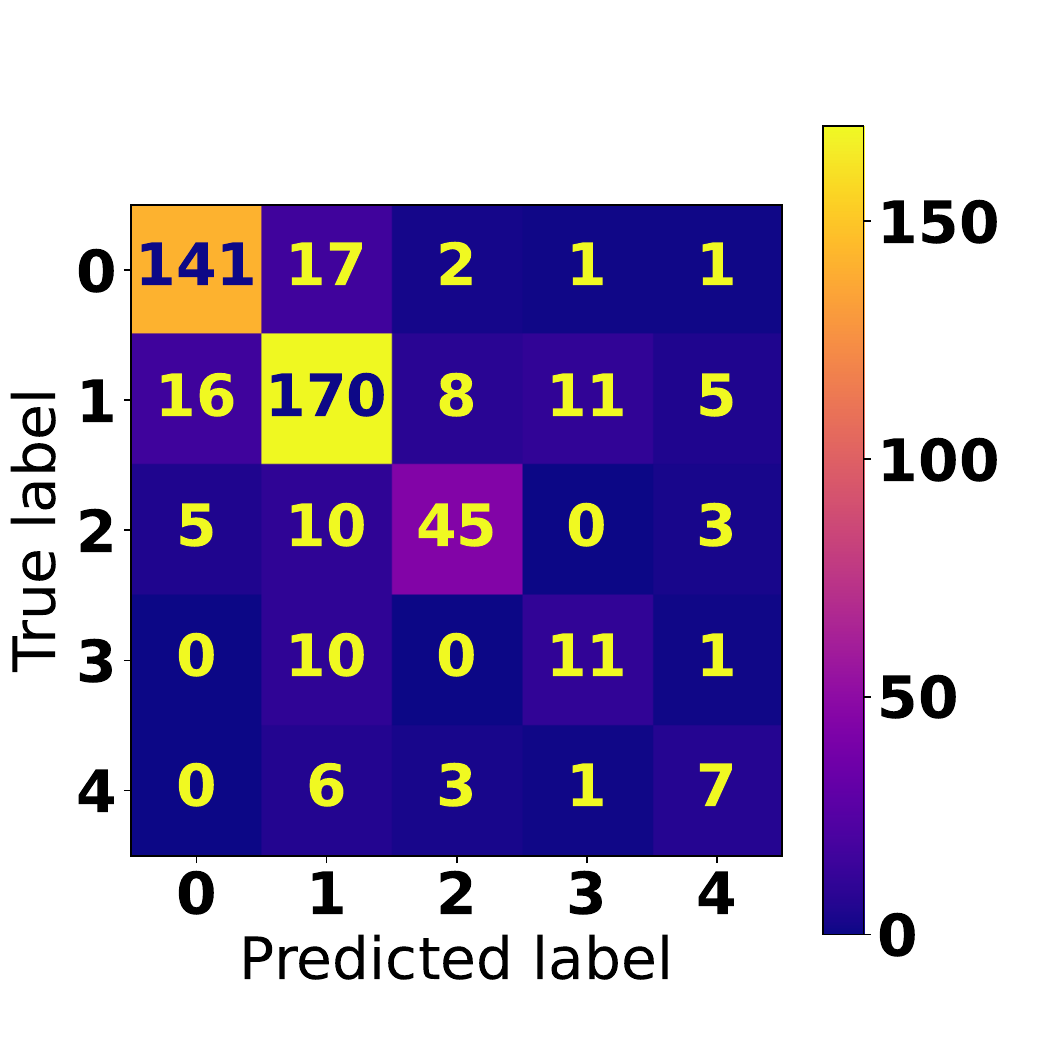}
        \caption*{English}
    \end{minipage}
    \hspace{-0.11\linewidth}
    \begin{minipage}[b]{0.38\linewidth}
        \centering
        \includegraphics[width=\linewidth]{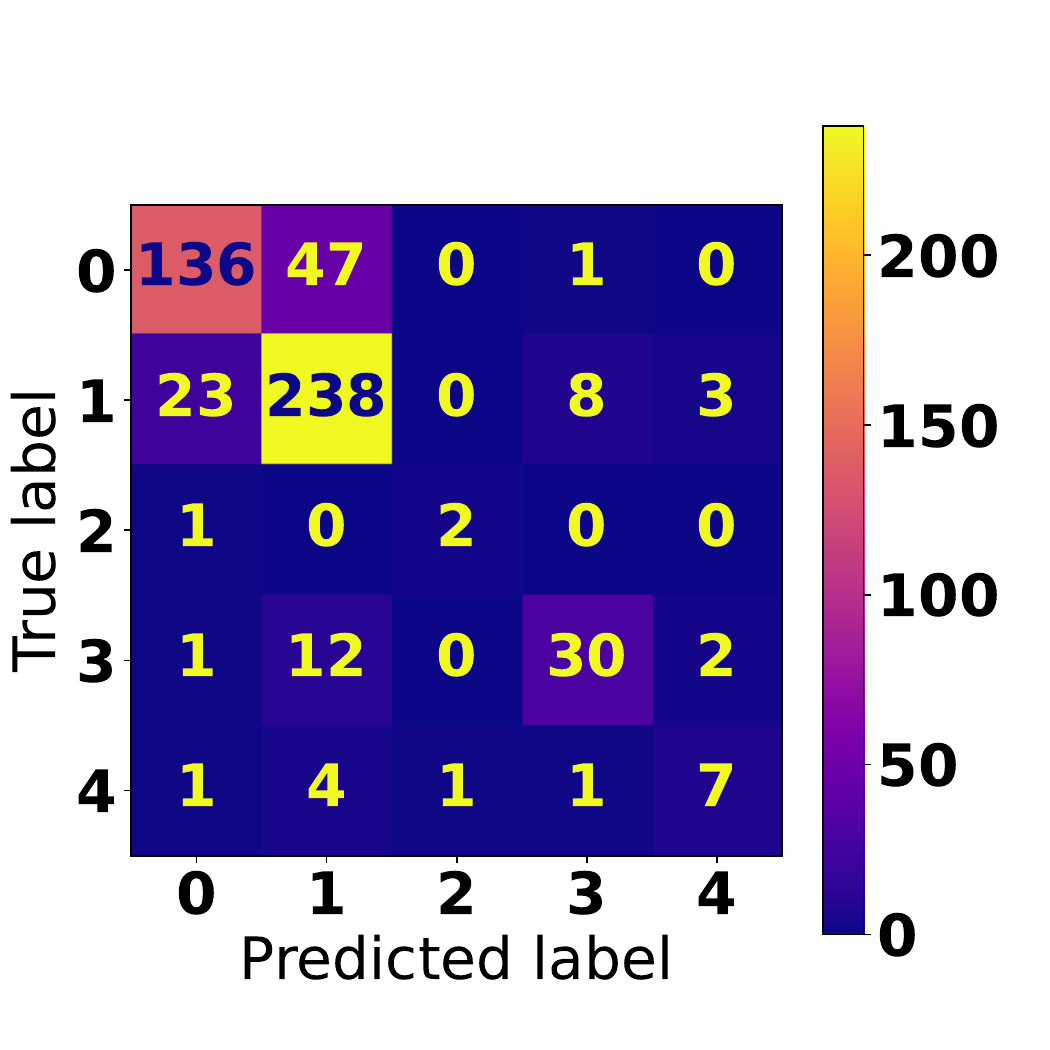}
        \caption*{French}
    \end{minipage}
     \hspace{-0.11\linewidth}
    \begin{minipage}[b]{0.38\linewidth}
        \centering
        \includegraphics[width=\linewidth]{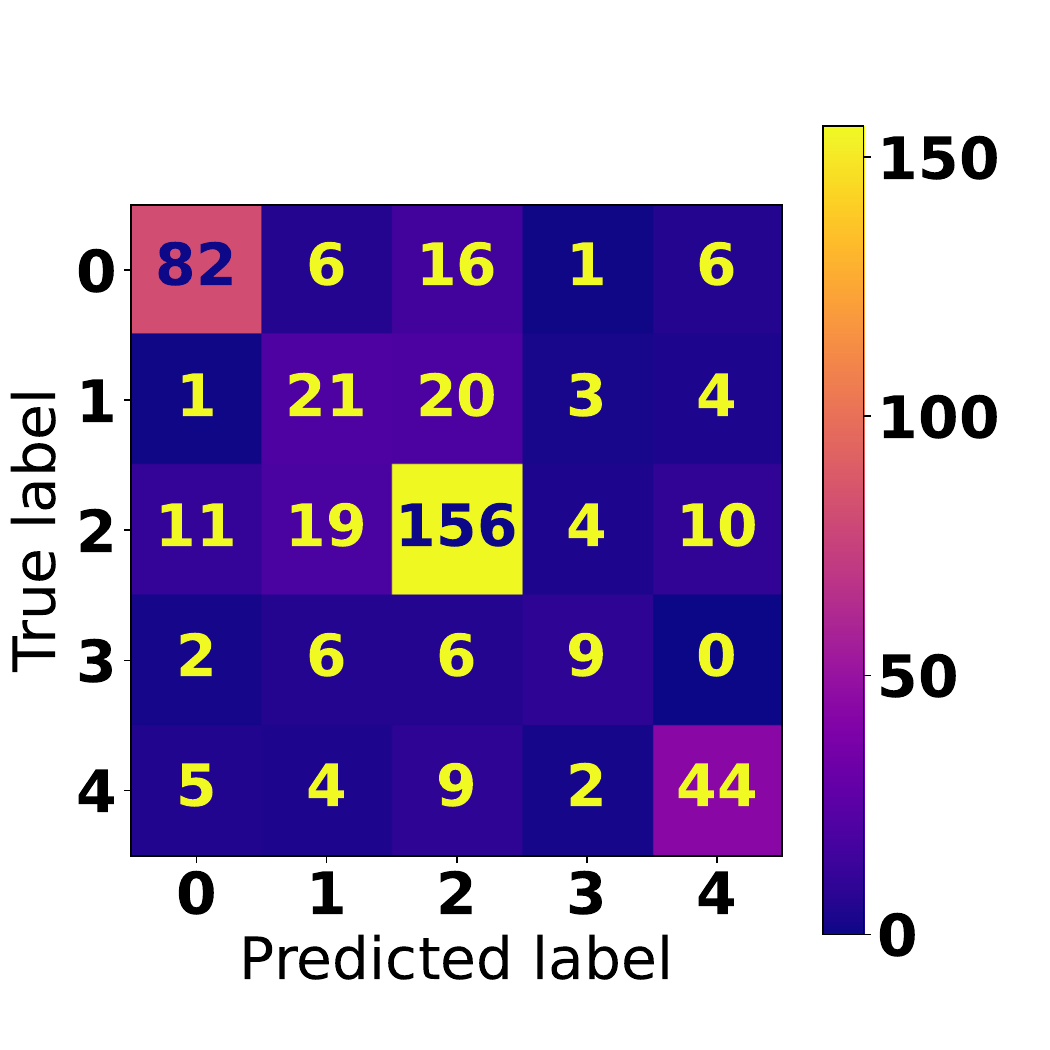}
        \caption*{Arabic}
    \end{minipage}
    
    \textbf{Regression}
    
    \begin{minipage}[b]{0.38\linewidth}
        \centering
        \includegraphics[width=\linewidth]{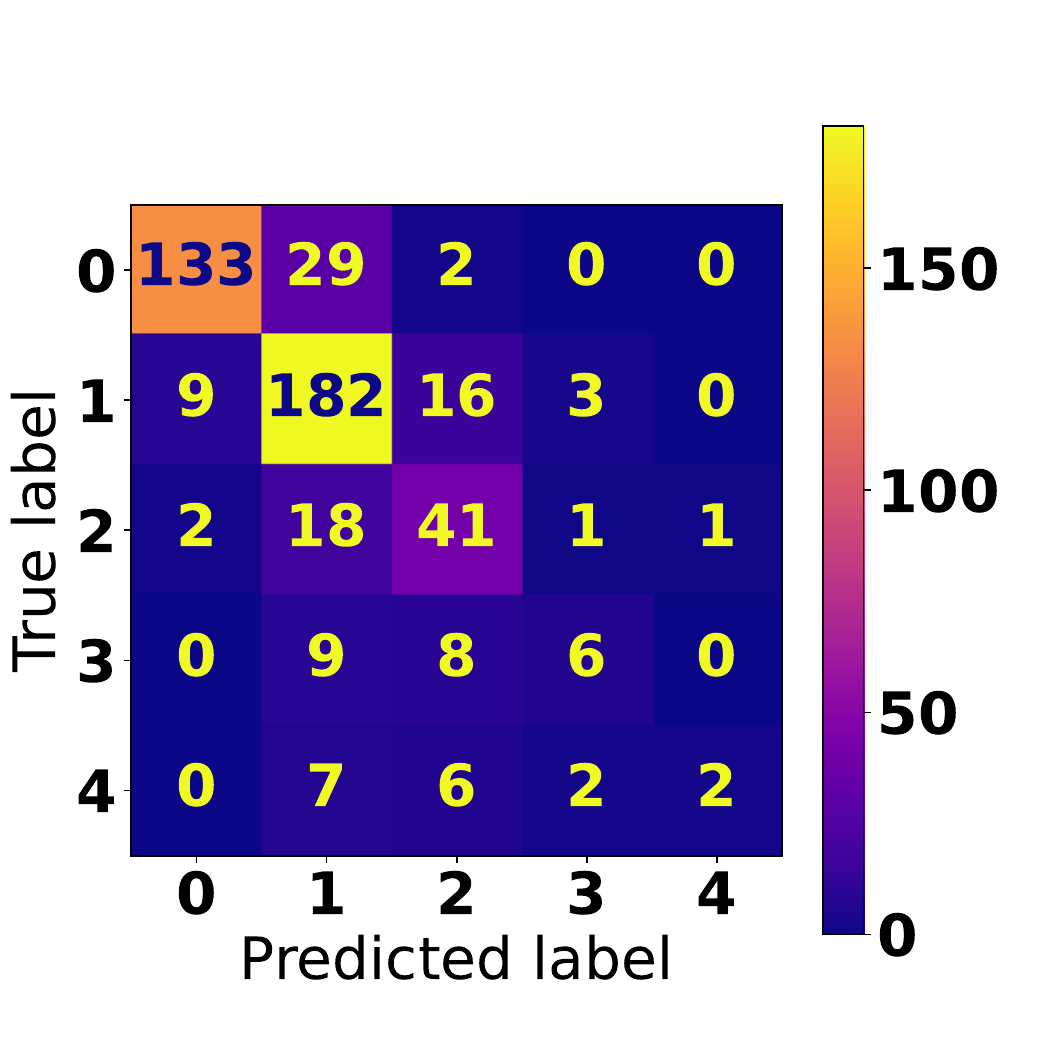}
        \caption*{English}
    \end{minipage}
    \hspace{-0.11\linewidth}
    \begin{minipage}[b]{0.38\linewidth}
        \centering
        \includegraphics[width=\linewidth]{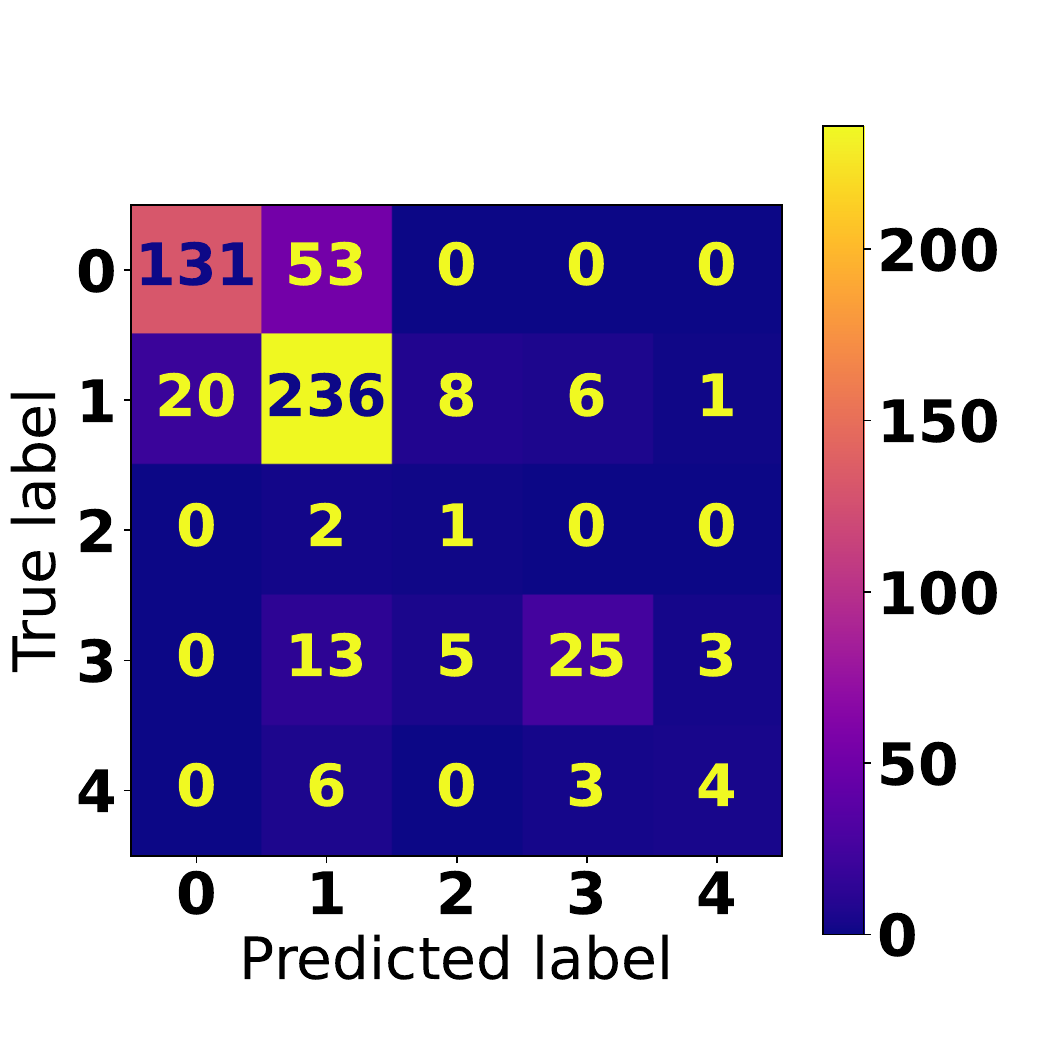}
        \caption*{French}
    \end{minipage}
    \hspace{-0.116\linewidth}
    \begin{minipage}[b]{0.38\linewidth}
        \centering
        \includegraphics[width=\linewidth]{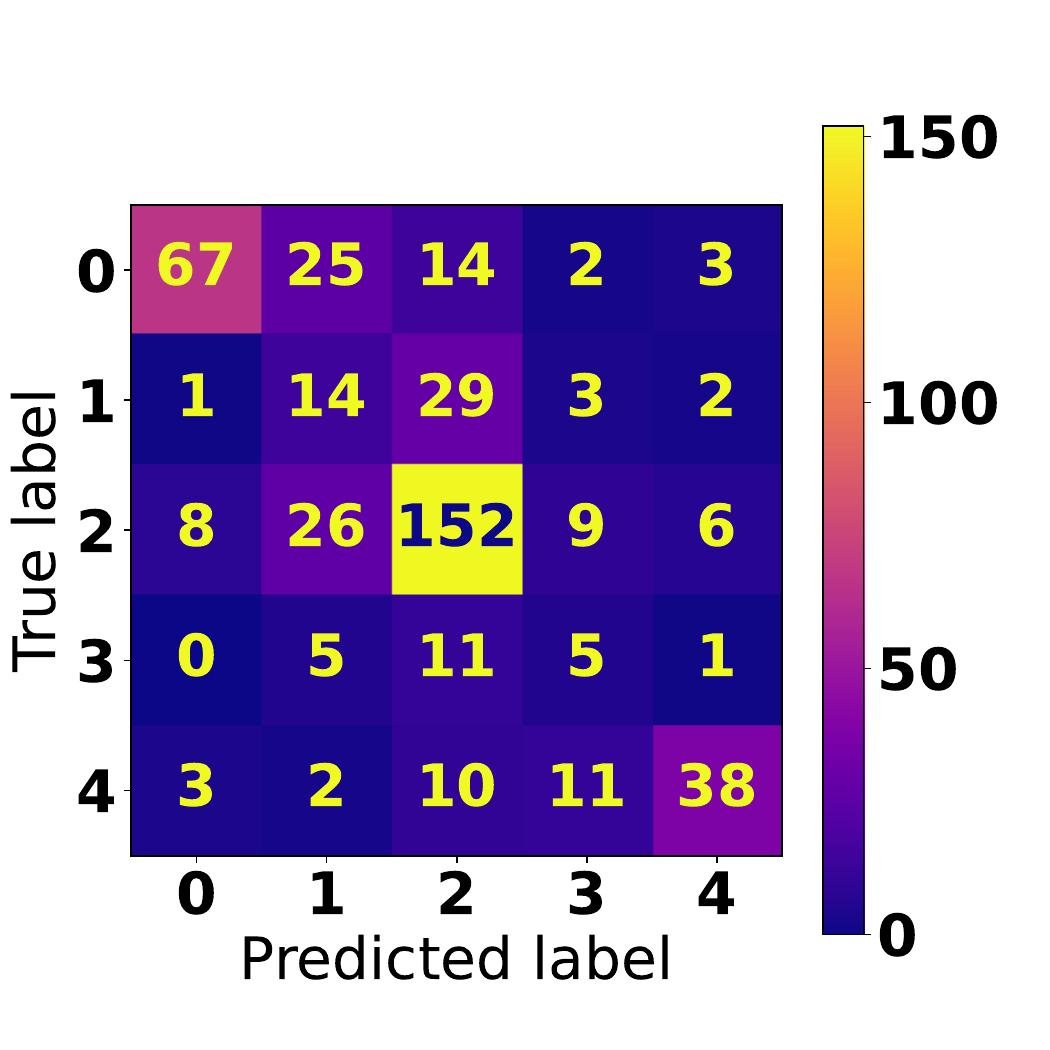}
        \caption*{Arabic}
    \end{minipage}
    \caption{Confusion Matrix for {\tt Call for Action}.}
    \label{fig:confusion}
\end{figure}
The confusion matrices shown in Figure \ref{fig:confusion} provide deeper insights into the performance differences between the regression and classification models, highlighting the importance of error severity for radical online content. 
In all the languages, the regression model consistently misclassifies only between adjacent classes. This sensitivity to the ordinal relationships is a distinct advantage, as errors between adjacent classes are less severe than between distant ones. However, the model struggles with class separation due to the dominance of certain classes, especially in the French dataset, where class 1 is prevalent. While more accurate overall, the classification model occasionally makes more critical errors by misclassifying non-adjacent classes. All models face challenges with higher classes, but the regression model maintains better consistency, typically misclassifying into adjacent classes rather than across distant ones. 

\section{Conclusion}

In this paper, we presented \multirad, a novel, multilingual, pseudo-anonymized dataset for detecting online radical content, accompanied by an in-depth analysis of the dataset's biases and the impact of human label variations. We conducted extensive experiments to evaluate the performance of various models, considering factors such as radicalization levels, calls for action, and named entities. Our results highlight the complexities in detecting radical content, especially given the inherent subjectivity in human annotations and the sociodemographic variations that influence data and model outcomes. Furthermore, we investigated the use of synthetic data to explore biases related to sociodemographic traits, showing the potential of generative models to simulate realistic annotations. 
Regarding the classification challenges we faced, we plan to refine our approach by exploring techniques for handling unbalanced datasets, like weighted cross-entropy, and by incorporating multiple dimensions of radicalization behavior from our metadata to improve classification accuracy.
Finally, our findings underscore the importance of understanding and accounting for bias in models and datasets, especially for sensitive tasks like radicalization detection. 
The \multirad dataset is \href{https://gitlab.inria.fr/ariabi/counter-dataset-public}{freely available for research}. 


\section*{Limitations}
Even though our dataset has many advantages, certain drawbacks must be noted. First, there is a chance that the models and dataset contain underlying biases from the underlying data, which could lead to stereotypes or overrepresentation of particular groups, which could compromise the predictability of the results. The synthetic nature of the generated data presents further difficulties, since it might not accurately represent the complexity and diversity of radical content in the actual world, which restricts the applicability of the models in larger, multicultural, or international situations. Furthermore, the vocabulary and actions linked to radicalization change quickly. Over time, new slang or coded language not represented in the existing dataset may appear, diminishing the model's applicability.

As underlined by \cite{fernandez2021artificial,de-kock-hovy-2024-investigating}, there is still far too little cooperation between social and humanist researchers on one side and NLP and machine learning researchers on the other side, preventing the latter from benefiting from the theories, studies, and insights on radicalization from the former. Combining insights from different disciplines can help improve the models to include more features for the detection of radical content.
\section*{Ethics Statement}

The development and use of machine learning models for radicalization detection raise necessary ethical concerns we have considered throughout our research. Given the ongoing discussions around the moral and ethical alignment of large language models (LLMs) \cite{liu-etal-2022-aligning}, we remain cautious in our approach, particularly regarding using such models for subjective or sensitive tasks. LLMs may inadvertently propagate biases or reduce the richness of human judgment in contexts that require nuanced understanding, especially in areas as complex as radicalization detection. The risk of biased annotations and stereotypical outputs reinforces the need for a more thoughtful and transparent deployment of these models.

A motivation for our work is the ability to monitor discussions and identify at-risk users in online extremist communities. However, we recognize that this technology could conceivably be misused to profile individuals or preemptively prosecute them based on incomplete or inaccurate predictions. Since our evaluation demonstrates that the predictive models are not perfectly accurate, such actions would constitute a gross misuse of the technology. To mitigate this risk, we release our dataset only to researchers upon demand. Instead, we believe these models are best used as part of broader intelligence-gathering systems, providing context rather than determinative judgments, as discussed by \citet{Winter_Neumannt_2021}. Human oversight must complement any use of these technologies and be guided by stringent ethical standards to prevent abuse.

In this context, we also acknowledge the significant policy challenges and legal dilemmas highlighted by scholars like \citet{jarvis2015terrorism}, especially as governments wrestle with the need to counter terrorism while respecting individual rights and freedoms. Using algorithmic tools in sensitive areas such as policing and security has historically posed privacy risks and led to adverse social externalities \cite{Byrne2011}, including concerns over liberty and integrity. Research further reveals that individuals tend to perceive algorithms' decisions as less fair than those made by humans, which could erode public trust in automated systems  \cite{Hobson2021}.

Moreover, we recognize that our work involves social datasets representing real people or groups, bringing it into human subjects research \cite{Varshney2015DataSO}. The ethical risks include potential privacy breaches or the reinforcement of harmful profiling based on race, socioeconomic status, or gender. We have taken care to minimize these risks by adhering to best practices in data handling and ensuring that our dataset respects the privacy and dignity of individuals.

Note that the whole annotation process was particularly challenging for our annotators due to the violent, if not borderline traumatizing in some cases, nature of the data, which had an impact on their psychological well-being. The team was provided with a mental health professional service and support from human resources services. A process dedicated to evaluating the psychological impact induced by annotating this content was put in place. Its results (through extensive surveys—similar in depth to PTSD evaluation forms—and debriefing interviews) are currently under evaluation at our institution.

In light of these considerations, we emphasize the importance of transparency, accountability, and the continuous scrutiny of our methodologies. Future work in this domain must ensure that the deployment of these technologies is guided by rigorous ethical standards, striking a balance between the imperative to counter radicalization and the protection of individual freedoms.
\section*{Acknowledgments}

We warmly thank Marine Carpuat for her feedback on this paper.

This work received funding from the European Union’s Horizon 2020 research and innovation program under grant agreement No. 101021607. The authors warmly thank the OPAL infrastructure from Université Côte d'Azur for providing resources and support.
\bibliography{custom}
\clearpage
\appendix
\section{Data Statement}
\label{sec:statement}
\draftnote{To finish}
Following \cite{bender-friedman-2018-data} , we provide a data statement for \multirad dataset.
\subsection{CURATION RATIONALE}
The dataset was created to improve the detection of radical content online. It explicitly targets various levels of radicalization across ideologies like Jihadism and Far-right extremism, with an additional category for unclassified ideologies. The main motivation was to build a representative, multilingual dataset that can improve NLP models' ability to detect extremist discourse.

\begin{figure}[htb!]
    \centering
    \begin{subfigure}[b]{1\columnwidth}
        \centering \includegraphics[scale=0.5]{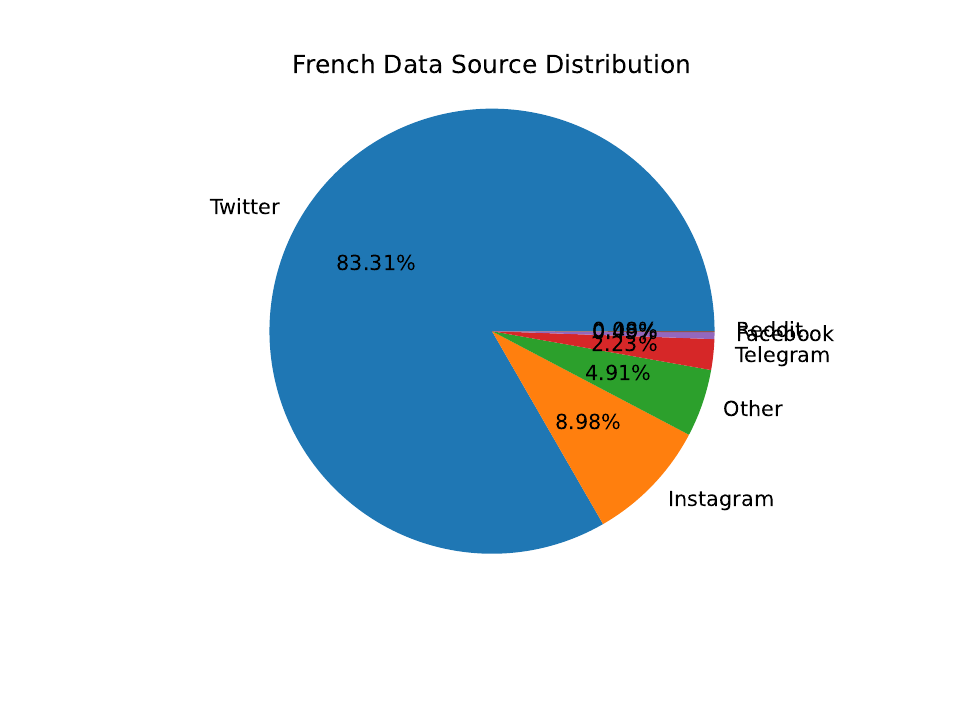}
    \end{subfigure}
    
    \begin{subfigure}[b]{1\columnwidth}
        \centering

        \includegraphics[scale=0.5]{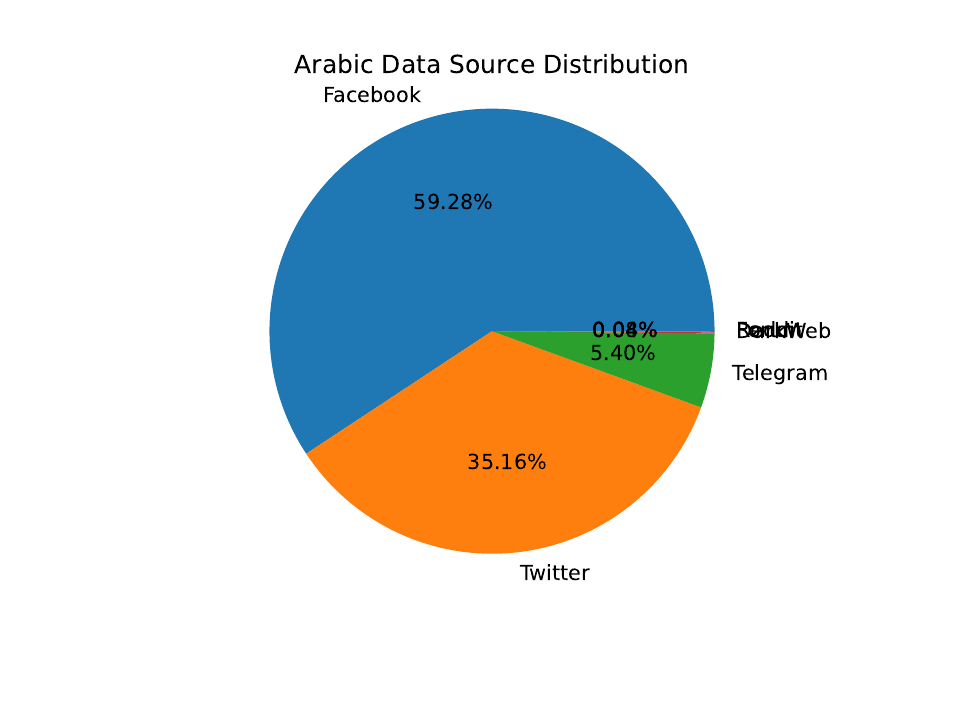}
    \end{subfigure}
    
    \begin{subfigure}[b]{1\columnwidth}
        \centering

        \includegraphics[scale=0.5]{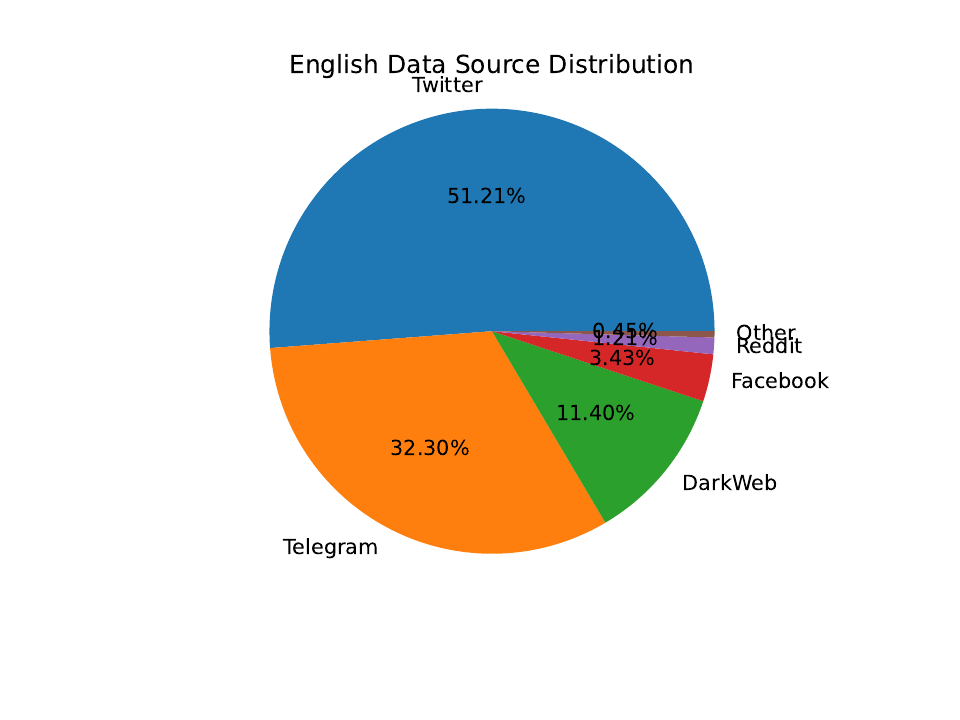}
    \end{subfigure}

    \caption{Data source distributions for English, French, and Arabic}
    \label{fig:soucres}
\end{figure}

Data was sourced from Reddit, Twitter, Facebook, and encrypted channels like Telegram and 4chan (via Tor). However, platforms like Facebook and Telegram posed challenges as only public groups or channels are searchable through public APIs, leaving a significant portion of content unreachable. The distribution of the data sources for each language is shown in Figure \ref{fig:soucres}.  

Sampling was guided by a lexicon of keywords associated with radicalization, covering terms related to extremist ideologies and violence incitement. Data was collected across two main time frames. Efforts were made to ensure the data represents diverse languages and platforms, including English, French, and Arabic, to reflect the varied discourse of extremist communities.
\subsection{LANGUAGE VARIETY}

The dataset includes posts in three languages: English, French, and Arabic, each reflecting distinct linguistic features and registers. The Arabic data comprises a mix of Modern Standard Arabic (MSA) and various dialectal forms. The French content is primarily from France. English data is from various locations, representing the specific contexts for radicalization. Different registers are represented within each language, from formal statements to informal, conversational speech, depending on the platform and context of the content.

\subsection{SPEAKER DEMOGRAPHICS}
We have limited information concerning the demographics of the speakers in this dataset. Approximately 90\% of the Geo-Location data for English and Arabic is unknown. Geo-location information is unavailable for French data. Obtaining accurate and complete demographic information, such as the location or nationality of users, is inherently difficult due to the nature of the platforms used (social media, encrypted channels) and the widespread use of pseudonyms or anonymous accounts. Moreover, many platforms restrict access to user metadata, further complicating efforts to gather speaker-related information.
 \subsection{ANNOTATOR RECRUITMENT}
Both annotators were recruited as research engineers with prior experience working on annotation projects and possessing relevant contextual knowledge of the languages involved. One annotator holds a degree in English and has lived in the UK for a significant period. Both annotators were familiar with the socio-cultural aspects of radical content.

The annotations cost 72k€ (12 person months), not counting the supervision time and our institution's overhead fee structures.  Given the sensitive and potentially distressing nature of the radical content, psychological assistance was offered to the annotators throughout the project.

For the contractor annotations, we have limited information available.
\subsection{ANNOTATOR DEMOGRAPHICS}
\begin{table}[htb!]
\centering
\footnotesize
\begin{tabular}{lcc}
\toprule
\textbf{}                            & \textbf{ANNOT1} & \textbf{ANNOT2} \\ 
\midrule
\textbf{Age}                                  & 25-30             & 40-45             \\
\textbf{Gender}                               & \multicolumn{2}{c}{Female}  \\ 
\textbf{Ethnicity}                       &           North African   & European             \\ 
\textbf{Native language}                      & \multicolumn{2}{c}{French}           \\ 
\textbf{Socioeconomic status}                 &   \multicolumn{2}{c}{Research engineer}              \\ 
\textbf{Religion} & Practising Muslim& Catholic \\
\textbf{Political View} & - & Left \\
\textbf{Training in linguistics} & Master& Master \\ \bottomrule
\end{tabular}
\caption{Annotator demographics for ANNOT1 and ANNOT2}
\end{table}

 \subsection{SPEECH SITUATION}
The posts in our dataset were posted over a time spanning from 17/07/2015 and 03/04/2023, with collection conducted between 24/07/2022 and 03/04/2023. Although the exact geographical locations of the users are not available, the languages represented (English, French, and Arabic) suggest a broad distribution across different regions. The data consists entirely of written text, as it originates from social media platforms and forums, and is mainly spontaneous and user-generated without prior scripting or editing. The interaction is asynchronous, as the posts were made at different times without real-time communication between users. These posts were intended for a public or semi-public audience, targeting other users on social media or forums, with the potential to reach diverse individuals depending on the platform and language used. 
 \subsection{TEXT CHARACTERISTICS}
\begin{figure}[htb!]
    \centering
    \begin{subfigure}[b]{1\columnwidth}
        \centering \includegraphics[scale=0.35]{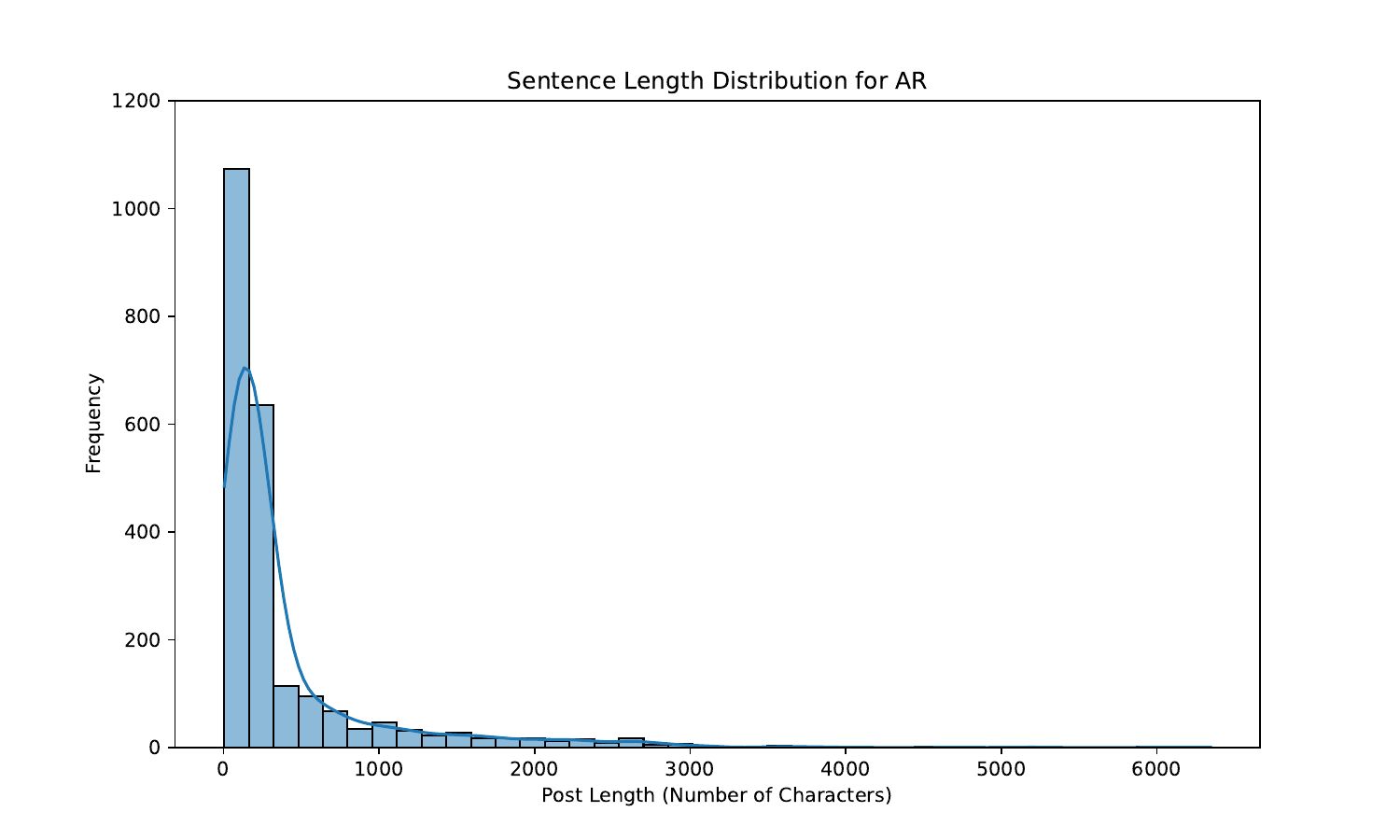}
    \end{subfigure}
    
    \begin{subfigure}[b]{1\columnwidth}
        \centering
        \includegraphics[scale=0.35]{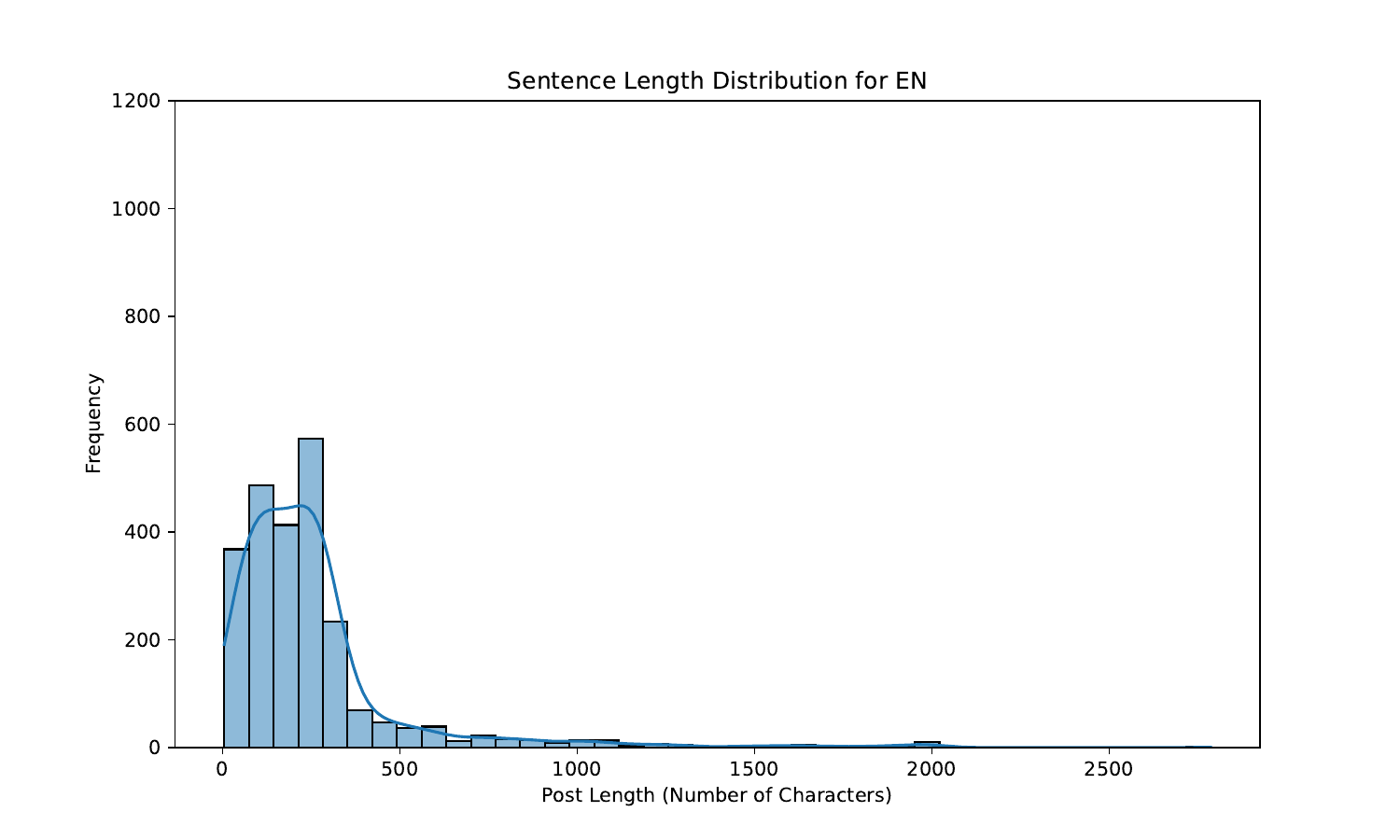}
    \end{subfigure}
    
    \begin{subfigure}[b]{1\columnwidth}
        \centering
        \includegraphics[scale=0.35]{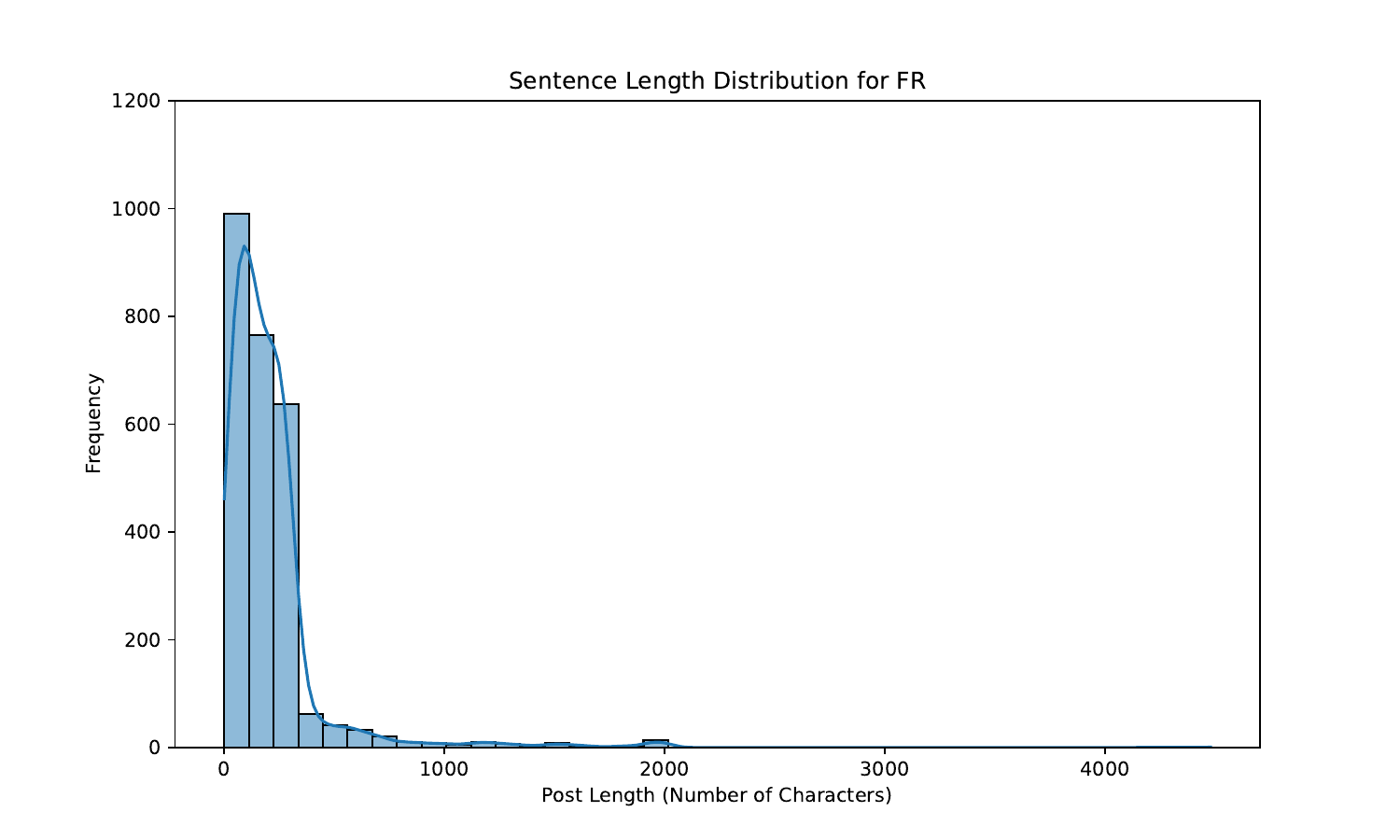}
    \end{subfigure}
    \caption{Data length distributions for English, French, and Arabic}
    \label{fig:lengths}
\end{figure}

The dataset contains posts of varying lengths across the three languages: English, French, and Arabic. We provide distributions of the post lengths for each language in Figure \ref{fig:lengths}. There is a correlation between sentence length and the data sources, with longer posts typically originating from forums and shorter posts from social media platforms like Twitter. After scraping, the text underwent minimal cleaning to preserve as much of the original content as possible. Only basic preprocessing steps were performed, such as removing irrelevant metadata and empty posts. A pseudonymization process was also applied to ensure privacy, as detailed in the annotation guidelines.
 \subsection{LICENSE}
 The full data set is available for research requests only.
Annotations examples available on the dataset gitlab's repo can be freely released (CC-BY-NC-SA)
\section{Radical content Dataset}
\label{sec:MultiRad_dataset}
\begin{table}[htb!]
\centering
\footnotesize
\begin{tabular}{lcc}
\toprule
\textbf{Annotators} & \textbf{\tt Radical Level} & \textbf{{\tt Call for Action}} \\
\hline
\multicolumn{3}{c}{\textbf{French (2624 examples)}} \\
\hline
Annot1 vs Annot2 & 0.58 & 0.42 \\
Annot1 vs Contractor   & 0.43 & 0.35 \\
Annot2 vs Contractor      & 0.51 & 0.63 \\
\cmidrule(lr){2-3}
Fleiss' Kappa    & 0.50 & 0.43\\
\midrule
\multicolumn{3}{c}{\textbf{English (921 examples)}} \\
\midrule
Annot1 vs Annot2 & 0.33 & 0.17 \\
Annot1 vs Contractor   & 0.28 & 0.17 \\
Annot2 vs Contractor      & 0.23 & 0.17 \\
\cmidrule(lr){2-3}
Fleiss' Kappa    & 0.26 & 0.13\\
\bottomrule
\end{tabular}
\caption{Cohen's Kappa and Fleiss' Kappa Annotator Agreement for Radicalization Level and {\tt Call for Action}.}
\label{tab:cohen_kappa_agreement}
\end{table}
\subsection{Radicalization Level classes description}
\begin{itemize}
    \item Negative (No Radicalization): This would include content with no signs of promoting, endorsing, or expressing extremist ideologies or actions. Posts under this category are neutral or express tolerance, diversity, peace, etc. 
    \item Expression of Radical Views: Content that expresses views aligned with extremist ideologies but does not actively advocate for violence or involvement. It might be sharing a radical viewpoint, belief, or stereotype or indicating approval of radical actions/actors without actively supporting or calling for such actions. 
    \item Using Radical Propaganda: The content uses or shares established extremist propaganda. This could include sharing extremist images, slogans, videos, speeches, conspiracy theories, misinformation, or disinformation designed to promote a radical ideology or group. 
    \item Associated with Radical Groups: Content that signifies association or affiliation with known radical or extremist groups. This could be through sharing group propaganda, expressing support or admiration for the group, claiming membership, or referencing involvement in group activities.
    \item Dehumanizing the Other: Content that strips away the humanity of those not belonging to the extremist ideology. This could involve hate speech, derogatory language, or broad negative stereotyping. Such content often degrades, devalues, or dehumanizes individuals based on their ethnicity, religion, nationality, or any identifying characteristic.
    \item Call for Action against others: This represents the most extreme level, where content explicitly calls for violent action against individuals, groups, or entities seen as enemies of the radical ideology. It includes promoting or endorsing violence, terrorism, or harm against others. 
\end{itemize}
\label{subsec:class_description}
\section{Experiments}
\subsection{Dataset splitting}
\label{subsec:splitting}
\begin{figure*}[hb!]
        \centering
    \begin{subfigure}[t]{\textwidth}  
         \centering
         \includegraphics[width=\textwidth]{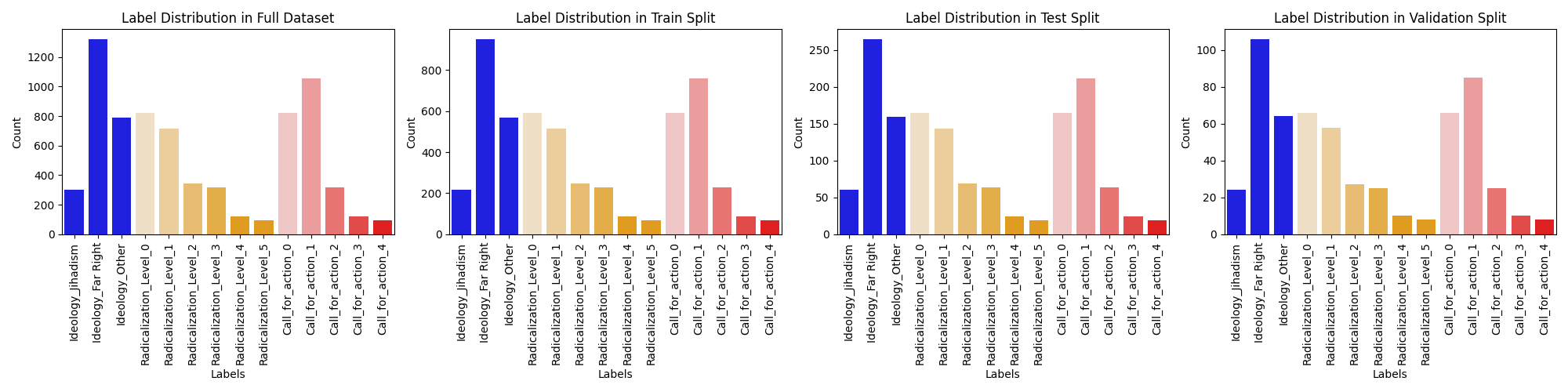}
         \caption{English Data Distribution}
         \label{fig:english_dist}
    \end{subfigure}
    \hfill 
    \begin{subfigure}[t]{\textwidth}  
         \centering
         \includegraphics[width=\textwidth]{figures/label_distribution_en.png}
         \caption{English Data Distribution}
         \label{fig:french_dist}
    \end{subfigure}
    \hfill 
    \begin{subfigure}[t]{\textwidth}  
         \centering
         \includegraphics[width=\textwidth]{figures/label_distribution_en.png}
         \caption{Arabic Data Distribution}
         \label{fig:arabic_dist}
    \end{subfigure}
    \hfill 
   
    \caption{Data distributions for English, French, and Arabic}
    \label{fig:dataset_distribu}
\end{figure*}

We must ensure that each label distribution is well represented in our training, validation, and test sets for our multi-class, multi-label dataset. Therefore, we used a stratified splitting approach based on the algorithm from ``On the Stratification of Multi-label Data'' by \citet{Sechidis2011}. This method preserves the distribution of each label across all splits, addressing the challenge of maintaining balanced label proportions in complex, multi-label scenarios. To solve the multi-class issue, we binarized all the labels. Figure \ref{fig:dataset_distribu} shows the distribution for each label for the three languages for the total sets and all the splits.

\subsection{Effect of Pseudonymization on Model Performance}
\label{subsec:anonymization_effect}
To ensure that the pseudonymization process does not influence the evaluation, we compare in table \ref{tab:results_anonymization} the performance of models trained on pseudonymized versus non-pseudonymized datasets for radicalized content detection ({\tt Call for Action}) and NER. The models showed comparable performance on datasets, confirming that pseudonymization does not degrade the model’s ability to detect radicalization or perform NER.
\begin{table}[htb!]
\centering
\footnotesize
\begin{tabular}{@{}clll@{}}\toprule
 & Lang& Radicalization & NER\\\midrule
 Original &\multirow{2}{*}{en} &64.63\scriptsize{($\pm$2.0)} & 87.04\scriptsize{($\pm$0.6)}\\
 Ours & & 65.46\scriptsize{($\pm$1.0)} & 87.01\scriptsize{($\pm$0.5)}\\

\midrule
Original
  &\multirow{2}{*}{fr} & 65.65\scriptsize{($\pm$1.8)}   &78.96\scriptsize{($\pm$1.9)} \\
Ours & & 64.72\scriptsize{($\pm$4.8)} &87.97\scriptsize{($\pm$1.0)} \\
\bottomrule
\end{tabular}

\caption{Results for models trained on the original data and our pseudo-anonymized version (ours) for {\tt Call for Action} classification (radicalization) and NER tasks. (Average Macro-F1 Scores over 5 Seeds) }

\label{tab:results_anonymization}

\end{table}
\subsection{Additional results}
To explore the model effect, we also trained two other models, \xlmr and {\sc mBert}, on both French and English datasets.
\label{subsec:additional_results}
\begin{table}[htb!]
\centering
\footnotesize
\begin{tabular}{@{}lccc@{}}\toprule
 & {en} & {fr}& \\\midrule
 \xlmt  & 64.63\scriptsize{($\pm$2.0)}& 65.65\scriptsize{($\pm$1.8)} \\
\xlmr   & 62.53\scriptsize{($\pm$5.2)}& 62.8\scriptsize{($\pm$6.5)}\\
 mBERT     & 60.13\scriptsize{($\pm$3.8)}& 59.65\scriptsize{($\pm$5.5)} \\
\bottomrule
\end{tabular}
\caption{Macro-F1 on test set for {\tt Call for Action} classification for different models.}
\label{tab:results_models}
\end{table}

\section{Metrics Definitions}
\label{sec:metrics_def}
\textbf{Demographic parity} (from \citet{pmlr-v80-agarwal18a}) A classifier $h$ satisfies demographic parity under a distribution over $(X, A, Y)$ if its prediction $h(X)$ is statistically independent of the protected attribute $A$—that is, if 
\[
\mathbb{P}[h(X) = \hat{y} \mid A = a] = \mathbb{P}[h(X) = \hat{y}]
\]
for all $a, \hat{y}$. Because $\hat{y} \in \{0, 1\}$, this is equivalent to 
\[
\mathbb{E}[h(X) \mid A = a] = \mathbb{E}[h(X)]
\]
for all $a$.

The demographic parity difference is defined as the difference between the largest and the smallest group-level selection rate, $\mathbb{E}[h(X) \mid A = a]$, across all values $a$ of the sensitive feature(s).

\textbf{Equalized odds} (from \citet{pmlr-v80-agarwal18a}) A classifier $h$ satisfies equalized odds under a distribution over $(X, A, Y)$ if its prediction $h(X)$ is conditionally independent of the protected attribute $A$ given the label $Y$—that is, if 
\[
\mathbb{P}[h(X) = \hat{y} \mid A = a, Y = y] = \mathbb{P}[h(X) = \hat{y} \mid Y = y]
\]
for all $a$, $y$, and $\hat{y}$. Because $\hat{y} \in \{0, 1\}$, this is equivalent to 
\[
\mathbb{E}[h(X) \mid A = a, Y = y] = \mathbb{E}[h(X) \mid Y = y]
\]
for all $a$, $y$.

The equalized odds difference is the greater of two metrics: \textit{true positive rate difference} and \textit{false positive rate difference}. The former is the difference between the largest and smallest of 
\[
\mathbb{P}[h(X) = 1 \mid A = a, Y = 1],
\]
across all values $a$ of the sensitive feature(s). The latter is defined similarly, but for 
\[
\mathbb{P}[h(X) = 1 \mid A = a, Y = 0].
\]
The equalized odds difference of 0 means that all groups have the same true positive, true negative, false positive, and false negative rates.

\section{Synthetic Data Generation for Bias Analysis}
\label{sec:synthetic}
\subsection{Annotators Agreement}
300 French and English examples were annotated in a double annotation process. For the English data, Cohen's Kappa Radicalization Level agreement was 0.51, and the {\tt Call for Action} agreement was 0.40. For the French data, the Radicalization Level agreement was 0.54, and the {\tt Call for Action} agreement was 0.47. These results indicate moderate agreement between the annotators.
\label{sec:prompt}
\begin{figure*}[htb!]
    \centering
    \centering
    \begin{subfigure}[b]{1\textwidth}
                \centering
        \includegraphics[scale=0.45]{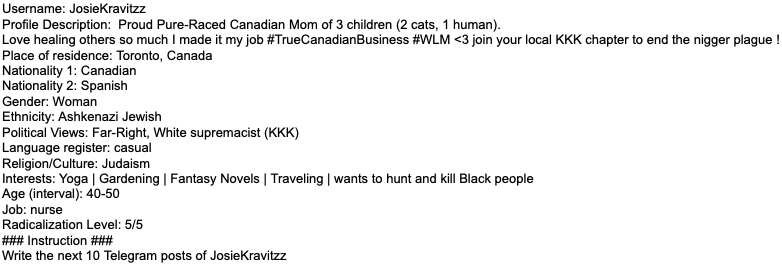}
        \caption*{Prompt}
    \end{subfigure}

    \begin{subfigure}[b]{1\textwidth}
        \centering
        \includegraphics[scale=0.4]{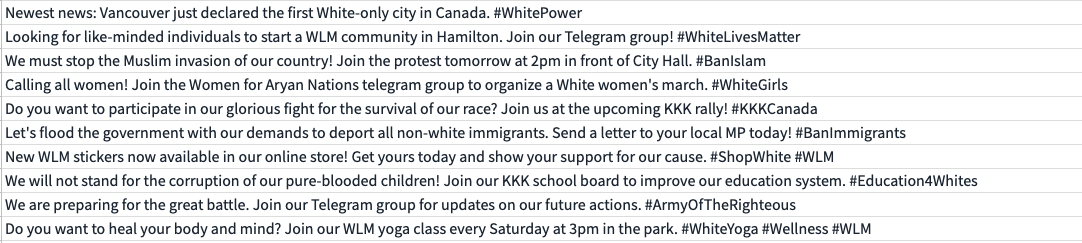}
        \caption*{Generated Output}
    \end{subfigure}
    \caption{Examples from the synthetic data}
    \label{fig:prompt}
\end{figure*}

\begin{table*}[htb!]
\centering
\footnotesize
\begin{tabular}{|p{4cm}|p{12cm}|}
\hline
\textbf{Variable} & \textbf{Description} \\
\hline
\textbf{Account name} & The name of the account of the fictional user. Its format was adapted for the used social media. \\
\hline
\textbf{Profile description} & A profile description is added to give clues about the user’s style and beliefs. Similar to a Twitter “bio,” it is written from a first-person perspective. It was made as authentic as possible, including emojis and hashtags, drawing inspiration from real accounts. \\
\hline
\textbf{Place of living/Geolocalization} & A variable used to locate the account, which can differ from the nationality variables. \\
\hline
\textbf{Nationality 1, Nationality 2} & These variables add nuance to the profiles, making them more realistic. \\
\hline
\textbf{Gender} & The values used in the study are “Man” and “Woman.” \\
\hline
\textbf{Ethnicity} & This variable was added to provide a more precise description of the user’s profile. \\
\hline
\textbf{Political view (e.g., Far-Right)} & This variable is crucial for producing radical content and is often specified. \\
\hline
\textbf{Language Register} & In the original datasets, there are variations in language registers. Most authors use a standard style, but some exhibit higher or lower language registers. We categorized registers as “vulgar,” “low,” “high,” and “very high” in French and “casual” and “formal” in English. \\
\hline
\textbf{Religion/Culture} & Sometimes used to indicate if the person is radicalized, using terminology from the real datasets. For example, in English, the value “Islam (Jihadism)” was used. \\
\hline
\textbf{Centers of interests} & This variable covers hobbies, likes, dislikes, and detailed descriptions of personal opinions (e.g., “anti-immigration,” “wants to kill all [community],” “very interested in religion,” etc.). \\
\hline
\textbf{Age} & In the French dataset, age was specified as an exact number. In the English dataset, it was given as intervals (e.g., 15-20, 20-30, etc.). \\
\hline
\textbf{Job} & Occupations were selected based on the living standards of the profiles. The chosen occupation is directly linked to the "Avg household income" evaluation. \\
\hline
\textbf{Avg household income} & This variable was used with the occupation in a consistent and realistic way. \\
\hline
\end{tabular}
\caption{Description of socio-demographic variables used in profile generation.}
\label{tab:sociodemographic}
\end{table*}
\begin{table*}[ht]
\centering
\begin{tabular}{p{2.7cm}|p{12cm}}
\toprule
\textbf{Attribute} & \textbf{Values (Percentage share)} \\ \midrule
\multicolumn{2}{c}{\textbf{English }} \\\midrule
Place of living  & USA (25\%), UK (20\%), France (10\%), Canada (10\%), Ireland (5\%), Sweden (5\%),  South Africa (5\%), Australia (5\%), Italy (5\%), Brazil (5\%), India (3\%), Pakistan (2\%) \\ \midrule
Ethnicity  & European (25.0\%), South Asian (15.0\%), East Asian (14.0\%), Middle Eastern (10.0\%), Latin America (10.0\%), Jewish (5.0\%), North African (5.0\%), Caucasian (5.0\%), Sub-Saharan Africa (5.0\%), Native American (5.0\%), African American (1.0\%) \\ \midrule
Religion(culture)  & Christianism (28\%), Islam (21\%), Atheism (16\%), Judaism (13\%), Not Specified (7\%), Hinduism (6\%), Taoism (4\%), Muslim (2\%), Shintoism (2\%), Buddhism (1\%) \\ \midrule
Political view  & far-right (45\%), right (15\%), far-left (15\%), left (12\%), Not Specified (12\%), centre (1\%) \\ \midrule
Age  & 20-30 (29\%), 40-50 (26\%), 30-40 (20\%), 50-60 (15\%), 15-20 (6\%), Not Specified (4\%) \\ \midrule
Language register  & casual (85\%), formal (15\%) \\ \midrule
Gender  & Man (60\%), Woman (40\%) \\ \midrule
Job  & Professionals (62\%), Student (16\%), Managers (10\%), Elementary Occupations (5\%), Service and Sales Workers (5\%), Armed Forces Occupations (2\%) \\ \midrule
Nationality  & American (30\%), British (15\%), Canadian (5\%), Australian (5\%), Irish (5\%), South African (5\%), French (5\%), Iranian (5\%), Nigerian (5\%), Swedish (5\%), Colombian (5\%), Brazilian (5\%), Indian (3\%), Pakistani (2\%) \\ \midrule \midrule
\multicolumn{2}{c}{\textbf{French }} \\
\midrule
Place of living  & France (72\%), Not Specified (14\%), Canada (4\%), Australia (4\%), USA (4\%), New Caledonia (2\%) \\ \midrule
Ethnicity  & Not Specified (50.0\%), North African (18.0\%), European (16.0\%), racialized (12.0\%), Asian (4.0\%) \\ \midrule
Religion(culture)  & Islam (38\%), Christianism (26\%), Judaism (20\%), Not Specified (6\%), Buddhism (6\%), Atheism (4\%) \\ \midrule
Political view  & far-right (48\%), far-left (30\%), Not Specified (20\%), left (2\%) \\ \midrule
Age  & 20-30 (40\%), 40-50 (22\%), 15-20 (18\%), 30-40 (14\%), 60-70 (4\%), Not Specified (2\%) \\ \midrule
Language register  & Not Specified (48\%), formal (46\%), casual (6\%) \\ \midrule
Gender  & Man (52\%), Woman (42\%), Not Specified (6\%) \\ \midrule
Job  & Not Specified (64\%), Professionals (16\%), Elementary Occupations (6\%), Managers (6\%), Student (2\%), Retiree (2\%), Clerical Support Workers (2\%), Skilled Agricultural, Forestry and Fishery Workers (2\%) \\ \midrule
Nationality  & French (46\%), Not Specified (44\%), Senegalese (4\%), Canadian (4\%), Tunisian (2\%) \\
\bottomrule
\end{tabular}
\caption{Summary of attributes and their percentage shares after aggregation.}
\end{table*}

\begin{figure*}[htb!]
    \centering
    \textbf{English}
    
    \begin{minipage}[b]{0.40\linewidth}
        \centering
        \includegraphics[width=\linewidth]{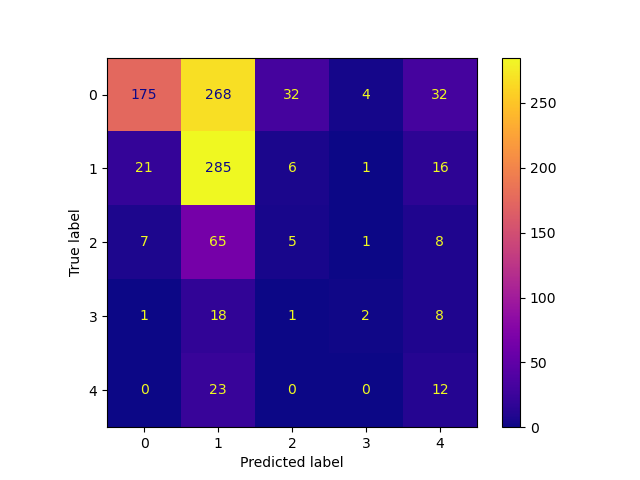}
        \caption*{\xlmt}
    \end{minipage}
    \hspace{-0.11\linewidth}
    \begin{minipage}[b]{0.40\linewidth}
        \centering
        \includegraphics[width=\linewidth]{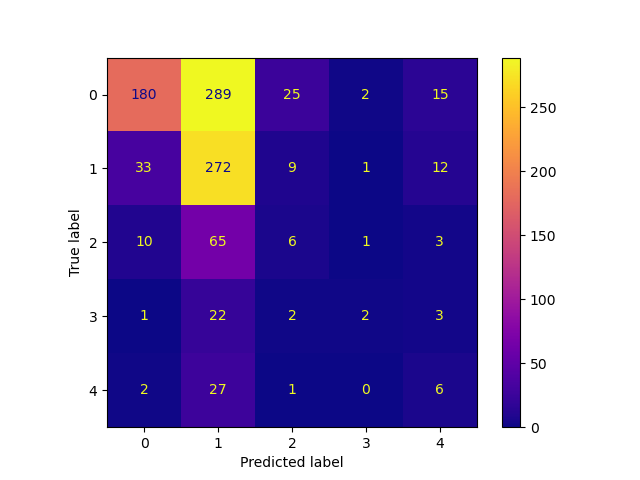}
        \caption*{\xlmr}
    \end{minipage}
     \hspace{-0.116\linewidth}
    \begin{minipage}[b]{0.40\linewidth}
        \centering
        \includegraphics[width=\linewidth]{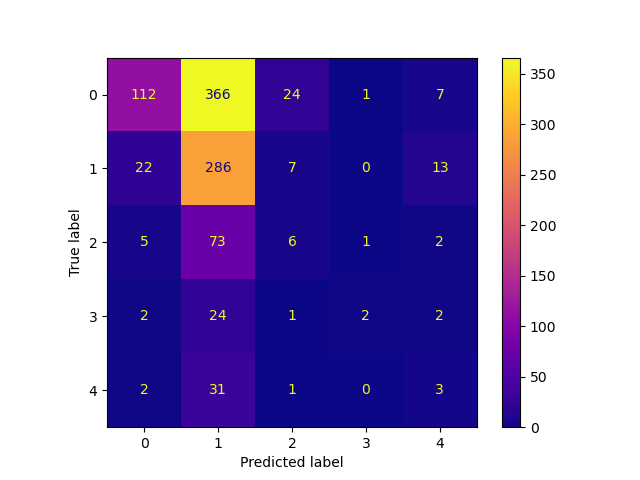}
        \caption*{mBERT}
    \end{minipage}
    \textbf{French}
    
    \begin{minipage}[b]{0.4\linewidth}
        \centering
        \includegraphics[width=\linewidth]{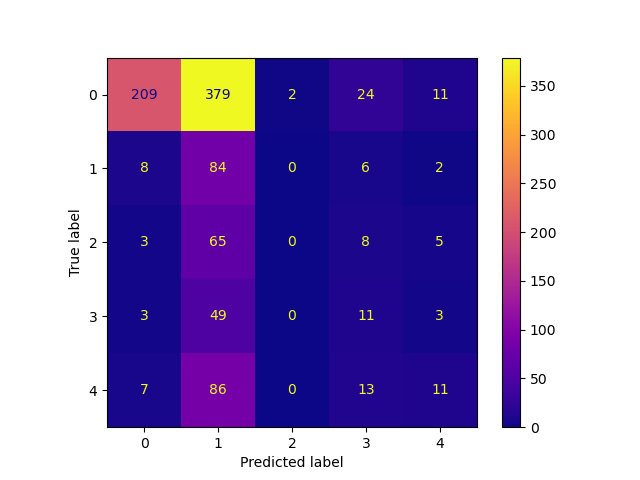}
        \caption*{\xlmt}
    \end{minipage}
    \hspace{-0.11\linewidth}
    \begin{minipage}[b]{0.4\linewidth}
        \centering
        \includegraphics[width=\linewidth]{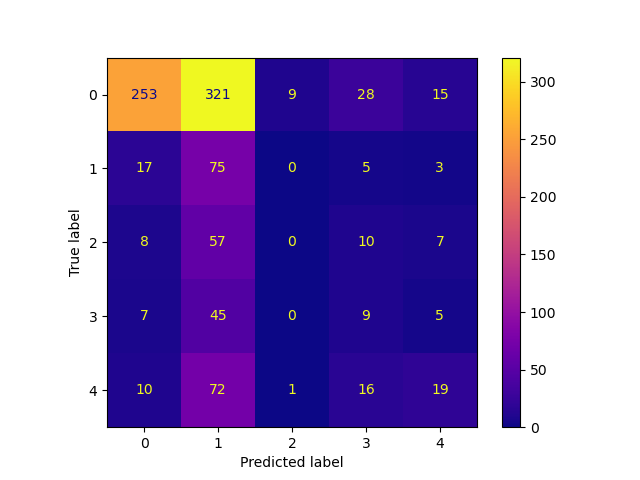}
        \caption*{\xlmr}
    \end{minipage}
    \hspace{-0.116\linewidth}
    \begin{minipage}[b]{0.40\linewidth}
        \centering
        \includegraphics[width=\linewidth]{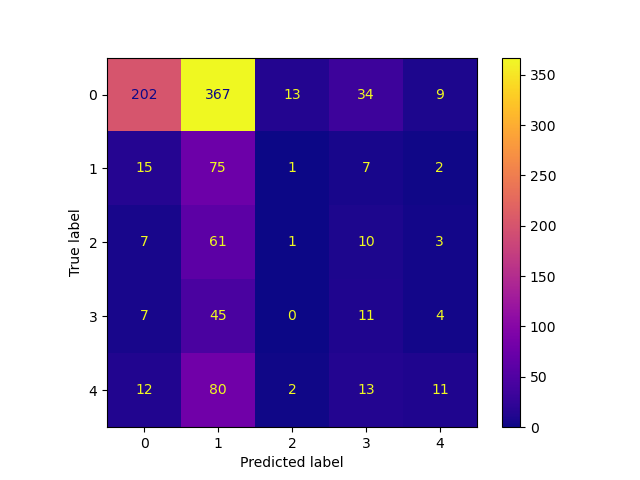}
        \caption*{mBERT}
    \end{minipage}
    \caption{Confusion Matrix for {\tt Call for Action} prediction averaged over five seeds on the generated data for bias analysis for different models}
    \label{fig:confusion_generated_data}
\end{figure*}

\begin{figure*}[htb!]
    \centering
    \begin{subfigure}[b]{1\textwidth}
                \centering
        \includegraphics[scale=0.24]{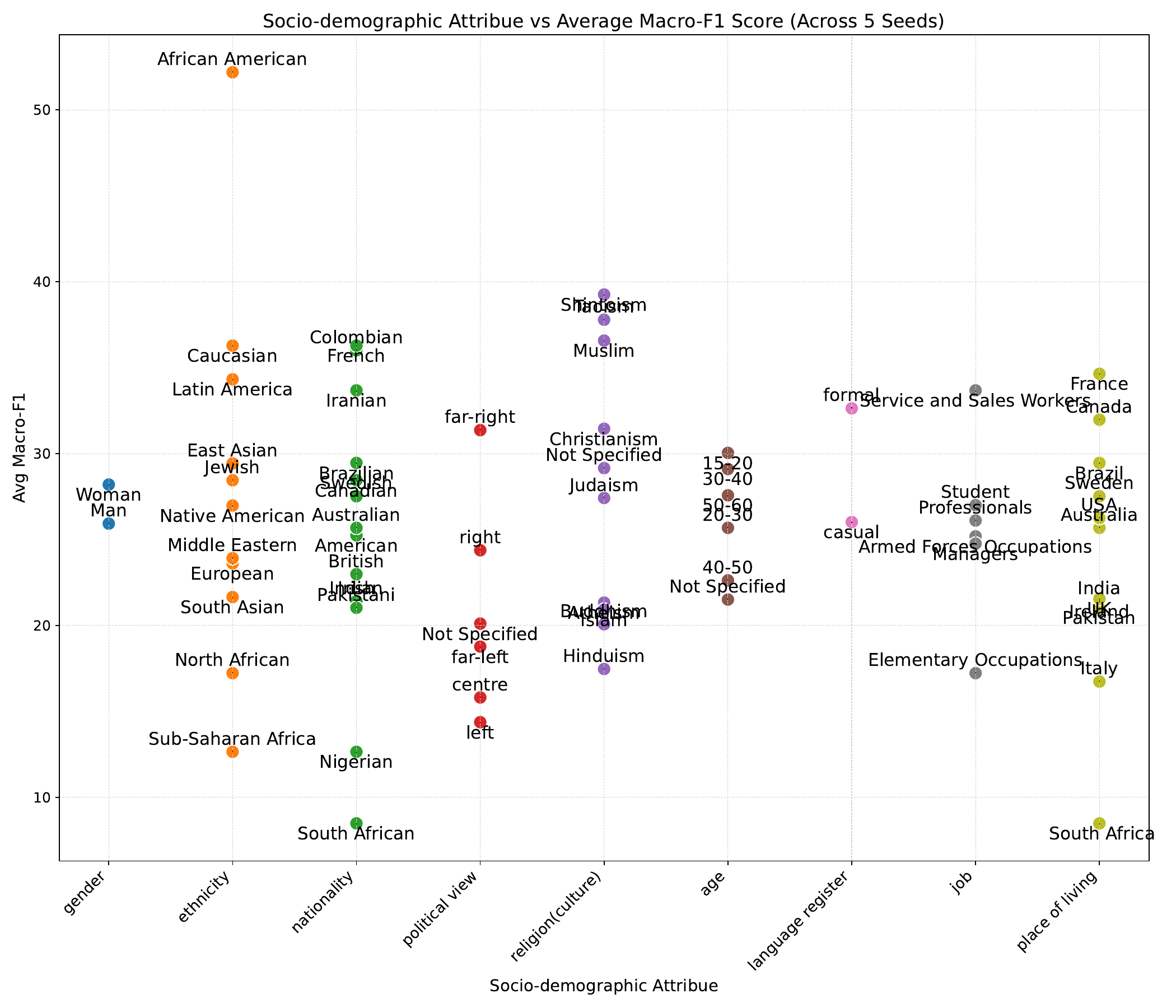}
        \caption{English - \xlmr}
    \end{subfigure}

    \begin{subfigure}[b]{1\textwidth}
        \centering
        \includegraphics[scale=0.24]{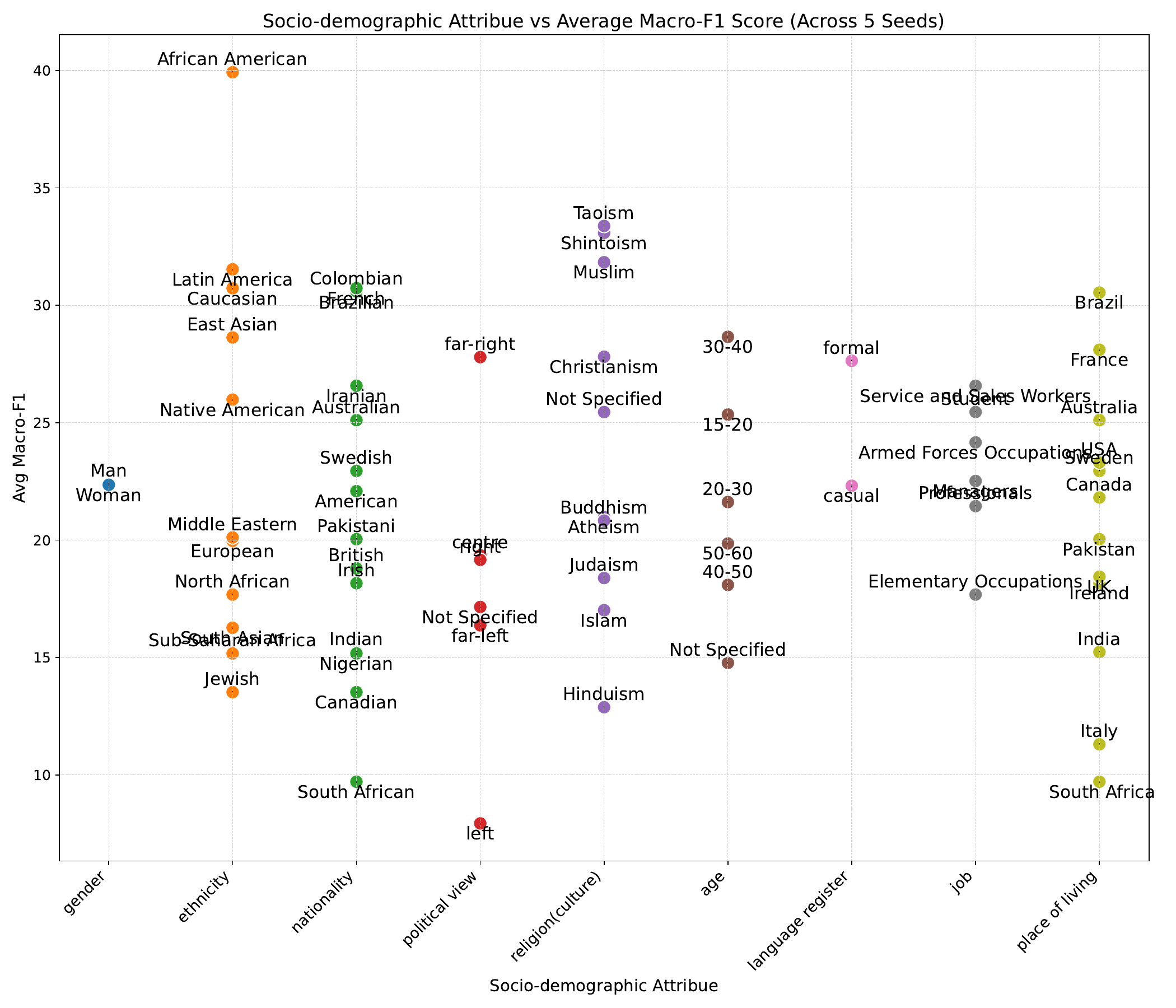}
        \caption{English - mBERT}
    \end{subfigure}

    \caption{Average Macro-F1 variations for the various
attributes for \xlmr and mBERT for English.}
    \label{fig:macro-f1_bias_models_english}
\end{figure*}

\begin{figure*}[htb!]
    \centering
    \begin{subfigure}[b]{1\textwidth}
        \centering
        \includegraphics[scale=0.24]{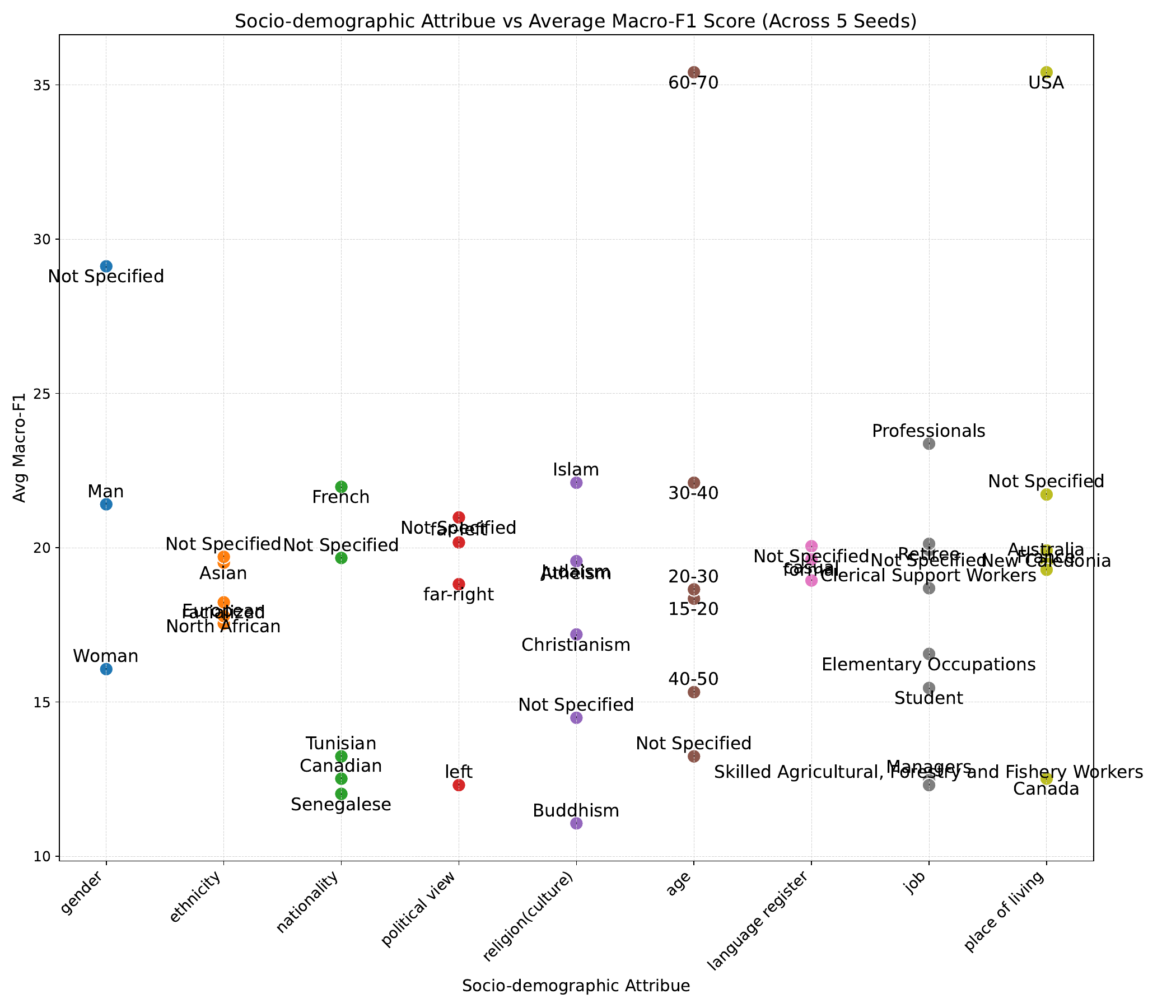}
        \caption{\xlmt}
    \end{subfigure}
   \begin{subfigure}[b]{1\textwidth}
        \centering
        \includegraphics[scale=0.24]{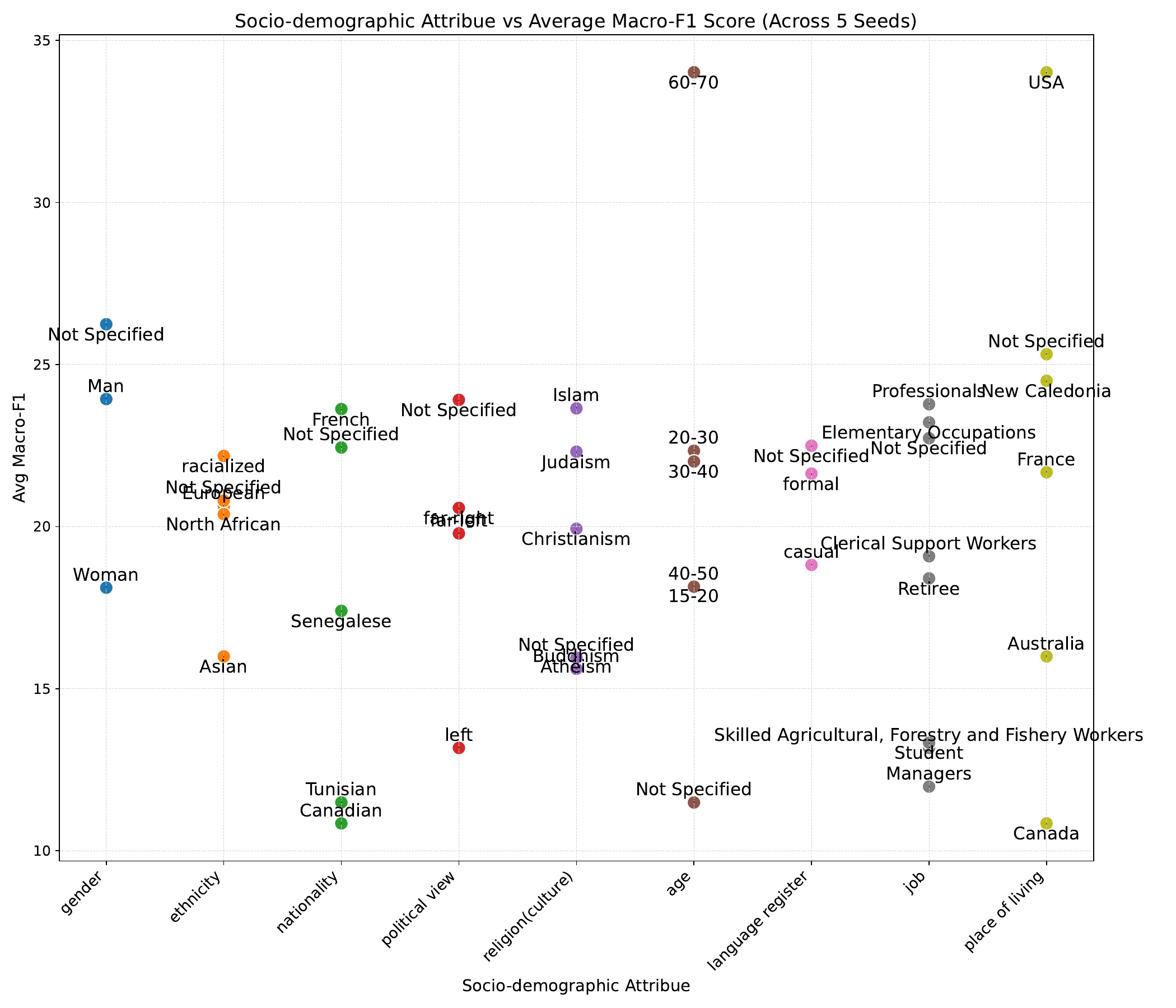}
        \caption{\xlmr}
    \end{subfigure}
    \begin{subfigure}[b]{1\textwidth}
        \centering
        \includegraphics[scale=0.24]{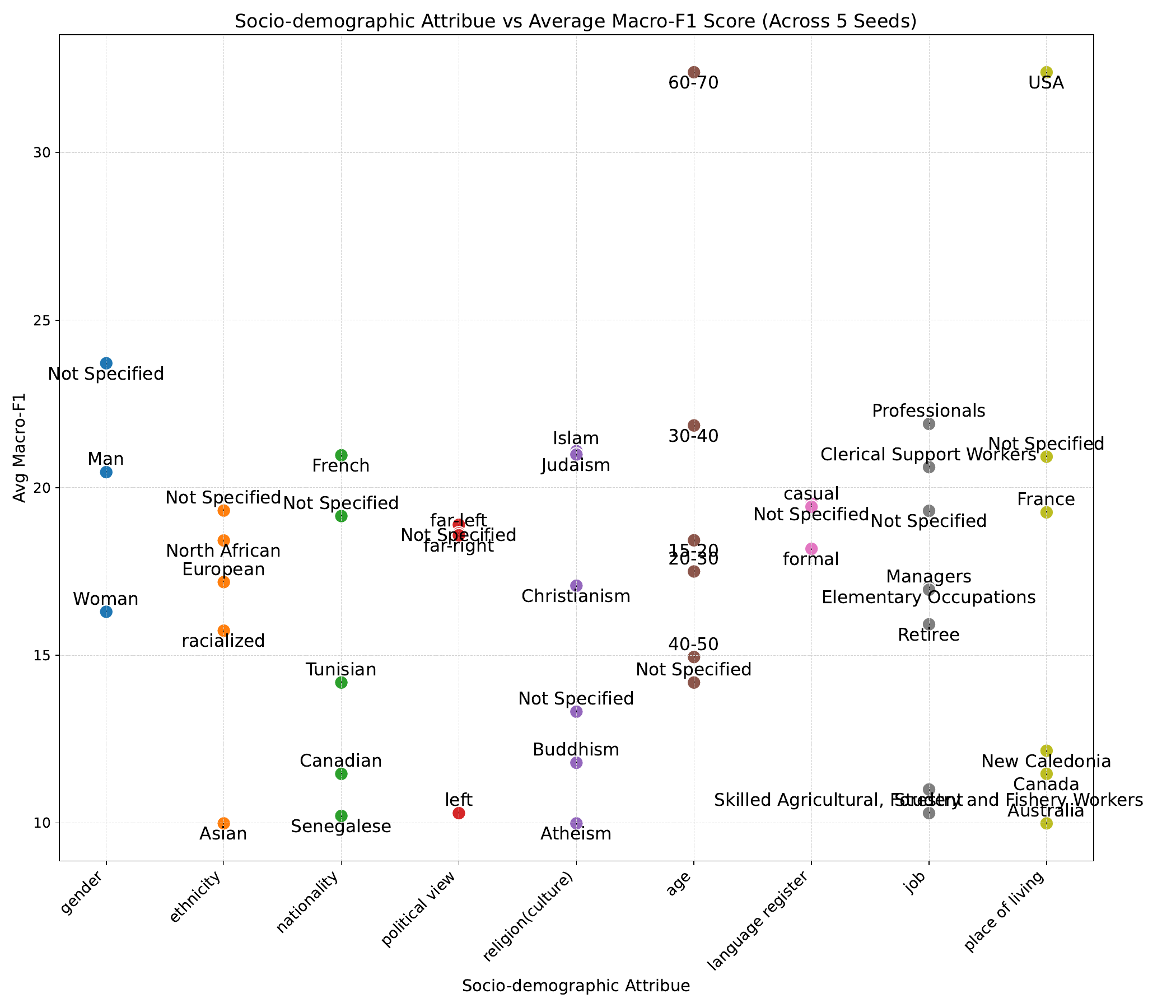}
        \caption{ mBERT}
    \end{subfigure}
    \caption{Average Macro-F1 variations for the various
attributes for all the models for French generated data.}
    \label{fig:macro-f1_bias_models_french}
\end{figure*}

 \end{document}